\documentclass{article}

\PassOptionsToPackage{numbers, sort}{natbib}

\usepackage[final]{neurips_2022}

\usepackage[utf8]{inputenc} 
\usepackage[T1]{fontenc}    
\usepackage{hyperref}       
\usepackage{url}            
\usepackage{booktabs}       
\usepackage{amsfonts}       
\usepackage{nicefrac}       
\usepackage{microtype}      
\usepackage{xcolor}         

\usepackage[ruled,vlined,linesnumbered]{algorithm2e}
\usepackage{algorithmic}

\usepackage[resetlabels]{multibib}
\newcites{SM}{Supplementary Material References}
\usepackage{enumitem}

\usepackage{amsthm}
\usepackage{shortcutsg}
\usepackage[capitalise]{cleveref}
\Crefname{assumption}{Assumption}{Assumptions}
\Crefname{section}{Section}{Sections}
\usepackage{etoolbox}
\AtBeginEnvironment{APPENDICES}{\crefalias{section}{appendix}}
\AtBeginEnvironment{APPENDICES}{\crefalias{subsection}{appendix}}

\theoremstyle{plain}
\newtheorem{theorem}{Theorem}

\theoremstyle{definition}

\newtheorem{remark}{Remark}

\newcommand{\IIDM}{{\mathrm{IIDM}}}
\newcommand{\FDIDM}{{\mathrm{FDIDM}}}

\title{The Implicit Delta Method}

\newcommand*\samethanks[1][\value{footnote}]{\footnotemark[#1]}

\author{%
  Nathan Kallus\thanks{Equal contribution, alphabetical order.} \\
  Cornell University\\ \& Netflix Research\\
  \texttt{kallus@cornell.edu}
  \And
  James McInerney\samethanks\\
  Netflix Research\\
  \texttt{jmcinerney@netflix.com}
}

\everypar\expandafter{\the\everypar\looseness=-1 }

\makeatletter
\def\thm@space@setup{%
  \thm@preskip=0.5\parskip \thm@postskip=-0.5\parskip
}
\makeatother
\usepackage[compact]{titlesec}  
\titlespacing\section{0pt}{0.2em plus 0.1em minus 0.1em}{0.2em plus 0.1em minus 0.1em}
\titlespacing\subsection{0pt}{0.2em plus 0.1em minus 0.1em}{0.2em plus 0.1em minus 0.1em}
\titlespacing\subsubsection{0pt}{0.2em plus 0.1em minus 0.1em}{0.2em plus 0.1em minus 0.1em}

\begin{document}

\maketitle

\begin{abstract}
Epistemic uncertainty quantification is a crucial part of drawing credible conclusions from predictive models, whether concerned about the prediction at a given point or any downstream evaluation that uses the model as input. When the predictive model is simple and its evaluation differentiable, this task is solved by the delta method, where we propagate the asymptotically-normal uncertainty in the predictive model through the evaluation to compute standard errors and Wald confidence intervals. However, this becomes difficult when the model and/or evaluation becomes more complex. Remedies include the bootstrap, but it can be computationally infeasible when training the model even once is costly. In this paper, we propose an alternative, the implicit delta method, which works by infinitesimally regularizing the training loss of the predictive model to automatically assess downstream uncertainty. We show that the change in the evaluation due to regularization is consistent for the asymptotic variance of the evaluation estimator, even when the infinitesimal change is approximated by a finite difference. This provides both a reliable quantification of uncertainty in terms of standard errors as well as permits the construction of calibrated confidence intervals. We discuss connections to other approaches to uncertainty quantification, both Bayesian and frequentist, and demonstrate our approach empirically.
\end{abstract}

\section{Introduction}

In this paper, we consider quantifying uncertainty in evaluations of predictive models trained on data.
Consider the following examples. 
We fit a complex model (such as a neural net) 
to predict mean service time for an incoming call 
to a call center given some features, 
and we use it to prioritize calls in a queuing system. 
We may be interested in confidence intervals 
on the average wait time of incoming calls in the queue.
Such confidence intervals would be crucial for drawing \emph{credible} conclusions about such evaluations, since we know we cannot take the point prediction at face value given the sampling uncertainty in the data.
We may, alternatively, be fitting a ranking algorithm by predicting user interaction from user-item features and then applying some fixed business rules on top, and we want to assess how often certain item categories would end up at the top. Of course, we would want to understand how certain we are in this assessment. 
Or, we fit a complex model to predict mean demand given price and user features from a price experiment we ran, and we use it to target discounts by optimizing demand at a price times unit profit. We may be interested in confidence intervals on the average profit over a given distribution of features.

All of these examples have three important features: they involve (1) a computationally burdensome step of fitting a large-scale model, (2) evaluating the result using a complicated function that need not even be known explicitly, and (3) requiring the epistemic uncertainty of the evaluation given a model and finite data set, 
in contrast to the total uncertainty comprising both epistemic and irreducible aleatoric uncertainty~\cite{der2009aleatory}. 
Were the first two of these simple (simple model and simple function thereof), we could just use the classic delta method \cite{doob1935limiting} (see next section for detail).
However, when these aspects are complex and the model involves many parameters, it may be too prohibitive to either analytically derive the whole inverse Fisher information matrix in the many model parameters or compute and invert the Hessian of the training loss empirically as well as compute the gradient of the final evaluation as a function of all parameters \cite{nilsen2022epistemic}. Even one aspect being complex may pose a serious challenge (\eg, uncertainty quantification for the prediction of a complex model at a point).
A remedy may be to bootstrap the whole process from data to final evaluation, but that can prove very computationally burdensome \cite{efron2021computer}.
Usually just fitting the model once is already an expensive task; fitting it hundreds of times can be operationally infeasible.
Other remedies, in the case of neural nets, may be the use of Langevin dynamics \cite{welling2011bayesian} or random dropout \cite{gal2016dropout}. But these assess uncertainty in network weights and/or network predictions, which must then be translated to uncertainty in the final evaluation.

In this paper, we propose a direct yet inexpensive way to generically assess uncertainty in such settings. Specifically, we consider conducting inference when the estimator is some specified function of a (conditional) maximum likelihood estimator (MLE), such as a regression or classification model. Our proposal, the \emph{implicit delta method}, works by simply adding an infinitesimal regularization to the MLE objective (\eg, the sum of squared errors). We prove that the infinitesimal change in the final estimator due to this regularization is consistent for its asymptotic variance, the same variance that would have been predicted by the delta method in theory. Hence, the name of our method: we are conducting a delta-method quantification of uncertainty implicitly without explicitly propagating the uncertainty through the derivative of the evaluation function, analytically deriving the possibly-huge Fisher information matrix, or approximating it empirically. We prove that even when we approximate the infinitesimal change with a finite difference with constant width, the change we measure is still consistent for the asymptotic variance. This not only gives an assessment of uncertainty in terms of standard errors, it also permits us to construct calibrated confidence intervals. We demonstrate this in experiments, comparing to other popular approaches for uncertainty quantification, both Bayesian and frequentist.

\begin{figure*}[t!]
\centering
\includegraphics[width=0.8\linewidth]{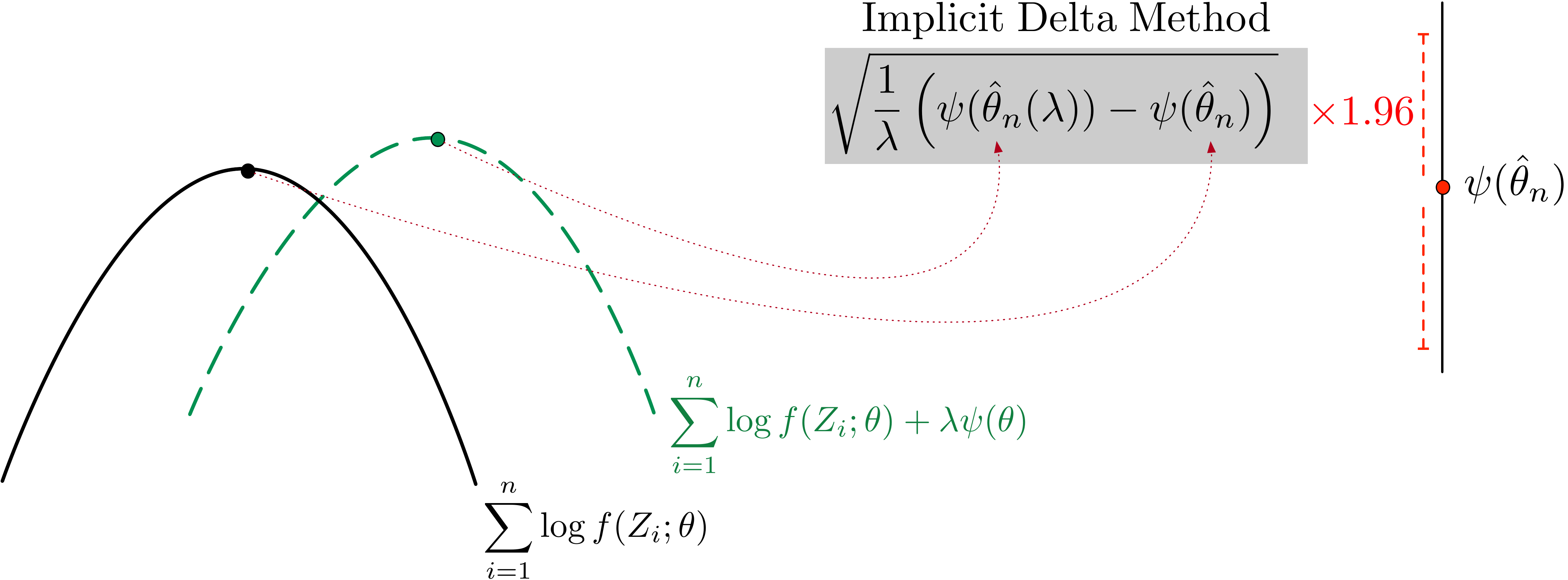}
\caption{Illustration of how the implicit delta method (IDM) estimates 95\% confidence intervals 
for a statistic of interest $\psi(\theta_0)$. Both the original MLE objective and the $\psi$-regularized objective are optimized, the $\psi$-evaluations of the two fitted models are compared, and $1.96$ multiples of the square root of the difference in evaluations is added and subtracted from the nominal evaluation to create a 95\% confidence interval.}
\labelfig{illustration_idm}
\end{figure*}

\section{Problem Set Up and the Delta Method}
\labelsec{setup_problem}

We consider an estimate constructed in two steps: first we fit a model using maximum likelihood estimation (MLE) and then apply some function to it.
Namely, we consider data given by $n$ independent and identically distributed (iid) observations $Z_i\in\Zcal$, $i=1,\dots,n$, drawn from a population with density $f(z;\theta_0)$ with respect to some base measure $\mu$ on $\Zcal$. For example, the data may consist of observations of features $X$ and responses $Y$, with $Z=(X,Y)$.

In the first step, we fit a model to the data by MLE. Given a model $\{f(z;\theta):\theta\in\Theta\}$ of densities (with respect to $\mu$) parametrized by $\theta\in\Theta\subseteq\R d$, we set
\begin{equation}\label{eq:mle}
\text{Model fitting:}\quad \hat\theta_n\in\argmax_{\theta\in\Theta}\sum_{i=1}^n\log f(Z_i;\theta).
\end{equation}
A prominent example is generalized regression, where we observe features and responses $Z=(X,Y)$, have a parametrized predictor $h_\theta(x)\in\R p$, and a parametric model $g(y;\vartheta)$ with $\vartheta\in\R p$. We then set $f((x,y);\theta)=g(y;h_\theta(x))$.\footnote{Note since we are not interested in the distribution of $X$ we here use only the conditional density of $Y\mid X$. Setting $f((x,y);\theta)=g(y;h_\theta(x))f(x)$ using the true unknown density $f(x)$ of $X$ does not change the MLE nor any of the results compared to omitting $f(x)$ altogether as we do here, which is referred to as the \emph{conditional} MLE by \citet{wooldridge2010econometric}.} Examples include least-squares regression, binary classification with cross-entropy loss, and Poisson regression, all with possibly complex and nonlinear predictors (\eg, neural nets).

In the second step, we process the trained model in some way to come up with our estimate. Given some $\psi:\Theta\mapsto\RR$, we compute
$$
\text{Evaluation using fitted model:}\quad \hat\psi_n=\psi(\hat\theta_n).
$$
One example in the case of generalized regression is evaluation of the predictor at a specified point, $\psi(\theta)=g_\theta(x_0)$. Another example is, when $g_\theta(x,p)$ corresponds to predicted mean demand at price $p$ given features $x$, we may be interested in the average optimal profit, $\psi(\theta)=\frac1m\sum_{j=1}^m \prns{\sup_{p\geq c}g_\theta(x_j,p)(p-c)}$, for an evaluation dataset $\{x_j:j=1,\dots,m\}$. More generally, $\psi$ could be more opaque: it could involve, for example, simulating a queuing system with a controller parametrized by $\theta$, such as a priority policy with priority score $g_\theta(x)$.

We are interested in conducting uncertainty quantification for $\hat\psi_n$, and in particular in inference on its population limit, $\psi_0=\psi(\theta_0)$.
One way to do this inference is to propagate through $\psi$ the uncertainty within $\hat\theta_n$ about $\theta_0$, provided we understand the latter uncertainty.
This is the so-called \emph{delta method}.

To apply it, we must first understand the uncertainty in $\hat\theta_n$. Provided some regularity holds, this uncertainty can be characterized by the \emph{curvature} of the objective function $\PP\ell(\theta,\cdot)$ at $\theta=\theta_0$:
if the curvature is sharp (resp., flat) then perturbing the objective and minimizing $\PP_n\ell(\theta,\cdot)$ instead does not (resp., does) move the minimizer far away. 
This curvature is exactly the \emph{Fisher information matrix}:
\begin{eqnarray}
I(\theta)=-\int (\nabla^2\log f(z;\theta))f(z;\theta)d\mu(z). \nonumber
\end{eqnarray}
Specifically, under appropriate regularity conditions,
\begin{align}
\label{eq:mleconv}
&\sqrt{n}(\hat\theta_n-\theta_0)\rightsquigarrow\Ncal(0,\,I^{-1}(\theta_0)).
\end{align}
In the above, $\Ncal(\mu,\Sigma)$ refers to the multivariate normal distribution and $\rightsquigarrow$ refers to convergence in distribution.
There are a variety of specific technical conditions that can establish this result. For an abstract presentation see theorems 9.27 and 9.28 in \citet{wasserman}. For more rigorous treatments see theorem 13.2 
of \citet{wooldridge2010econometric}, theorem 3.3 of \citet{neweymcfadden}, theorem 5.1 of \citet{lehmanncasella}, or theorem 8.3 of \citet{davidson1993estimation}, each of which uses slightly different technical regularity conditions.

Given \cref{eq:mleconv} holds with $I(\theta_0)\succ0$, the delta method would then guarantee that (see theorem 5.15 in \citealp{wasserman})
\begin{equation}\label{eq:deltamethod}
\sqrt{n}(\hat\psi_n-\psi_0)\rightsquigarrow\Ncal(0,\,V_0),\quad
V_0=\nabla\psi(\theta_0)\tr I^{-1}(\theta_0)\nabla\psi(\theta_0),
\end{equation}
provided $\nabla\psi(\theta_0)$ exists and $V_0>0$.

An immediate and very important consequence of this is that we can construct calibrated confidence intervals for $\psi_0$: under \cref{eq:deltamethod},
\begin{align}\label{eq:ci}
\text{if}~~~n\hat V_n\to_pV_0,~~~\text{then}~~~
&\Prb{\psi_0\in\bracks{\hat\psi_n\pm\Phi^{-1}((1+\beta)/2)\hat V_n^{1/2}}}\to\beta~~~\forall \beta\in(0,1).
\end{align}
where $\Phi$ refers to the cumulative distribution function of the standard normal distribution and $\to_p$ refers to convergence in probability.
For example, as long as $I(\theta)$ and $\nabla\psi(\theta)$ are continuous at $\theta_0$,
we can use
\begin{align}
\label{eq:Vn}
\hat V^{\mathrm{DeltaMethod}}_n=\frac1n\nabla\psi(\hat\theta_n)\tr
I^{-1}(\hat\theta_n)
\nabla\psi(\hat\theta_n).
\end{align}

As discussed in the introduction, however, this approach may prove intractable in practice, especially when $\theta$ has many dimensions. Since we are only truly concerned with the uncertainty in $\hat\psi_n$ and not in $\hat\theta_n$, it may seem unnecessary and overly cumbersome to first compute the uncertainty in the latter and then propagate it. We next present our method, which does this all implicitly, never working directly with the vector $\theta$ except as an optimization variable in maximizing the MLE objective and a perturbation thereof.

\section{The Implicit Delta Method}
\labelsec{idm}

We would like to construct calibrated confidence intervals as in \cref{eq:ci}, but computing the estimated standard error as in \cref{eq:Vn} can be prohibitive. The IDM is a way to compute the estimated standard error while \emph{neither} explicitly computing the uncertainty in $\hat\theta_n$ \emph{nor} propagating this uncertainty through $\psi$. Instead, we will simply slightly perturb the original MLE in \cref{eq:mle} using a little bit of regularization, which will implicitly do both of these difficult tasks for us.

To define the IDM, we first define a regularized version of the MLE. Given any $\lambda\geq0$, we consider adding the regularizer $\lambda\psi(\theta)$ to \cref{eq:mle} as well as the corresponding final estimator after passing through $\psi$:
\begin{equation}\label{eq:regmle}
\hat\theta_n(\lambda;\psi)\in \argmax_{\theta\in\Theta}\sum_{i=1}^n\log f(Z_i;\theta)+\lambda\psi(\theta),\quad\hat\psi_n(\lambda)=\psi(\hat\theta_n(\lambda;\psi)).
\end{equation}
We refer to this as $\psi$-regularized MLE.

We then define the \emph{infinitesimal} IDM (IIDM) as the infinitesimal change (\ie, derivative) in our final estimate using $\psi$-regularized MLE as we infinitesimally increase $\lambda$ from $0$:
\begin{equation}\label{eq:iidm}
\hat V_n^\IIDM=\frac{\partial}{\partial\lambda}\hat\psi_n(\lambda)\biggr\vert_{\lambda=0}=\lim_{\lambda\to0}\frac1\lambda\prns{\hat\psi_n(\lambda)-\hat\psi_n}.
\end{equation}

Our first result shows that the IIDM estimate is consistent for the true asymptotic variance in \cref{eq:deltamethod}.
\begin{theorem}\label{thm:iidm}
Suppose that $\hat\theta_n\to_p\theta_0\in\operatorname{Interior}(\Theta)$, $I(\theta_0)\succ0$,
and that, in a neighborhood of $\theta_0$, $\psi(\theta)$ is continuously differentiable and $f(Z;\theta)$ is almost surely twice continuously differentiable in $\theta$ with a Hessian that is bounded in operator norm by an integrable function of $Z$.
Then $$n\hat V_n^\IIDM\to_pV_0.$$
\end{theorem}
The significance of \cref{thm:iidm} is that, per \cref{eq:ci}, it implies that $\bracks{\hat\psi_n\pm\Phi^{-1}((1+\beta)/2)\sqrt{\hat V_n^\IIDM}}$ is a calibrated $\beta$-confidence interval for $\psi_0$.

Note that, aside from conditions on $\psi$ (which are the same as needed for \cref{eq:deltamethod,eq:Vn} to work), the regularity conditions required in \cref{thm:iidm} are implied by the regularity conditions required for establishing \cref{eq:mleconv} by, for example, any of \citet{neweymcfadden,lehmanncasella,wooldridge2010econometric,davidson1993estimation}. In that sense, these conditions are not strong as they are already needed for $\hat V_n^{\mathrm{DeltaMethod}}$ to be a good estimate of uncertainty to begin with, and they fit into the existing framework for the asymptotic analysis of MLE.

The implication of \cref{thm:iidm} is that we may be able to implicitly complete the steps of the delta method (compute the uncertainty in $\hat\theta_n$, then propagate it through $\psi$) by simply assessing the impact of regularizing the MLE. However, this requires we actually differentiate with respect to the regularization coefficient. While this requires computing just one first derivative (rather than many first and second derivatives as in \cref{eq:Vn}), it is still not clear how to do this in practice. 

In practice, we might approximate this derivative using finite differences, \ie, replace the limit in \cref{eq:iidm} with a very small $\lambda$.
This gives rise to what we call the finite-difference IDM (FDIDM), defined as follows for a given $\lambda_n>0$:
\begin{equation}\label{eq:fdidm}
\hat V_n^\FDIDM=\frac1{\lambda_n}\prns{\hat\psi_n(\lambda_n)-\hat\psi_n}.
\end{equation}

Our next result shows that it in fact suffices to choose $\lambda_n$ constant.
In fact any choice of $\lambda_n$ growing strictly slower than $n$, yields that $n\hat V_n^\FDIDM$ is also consistent for $V_0$, just like $\hat V_n^\IIDM$, provided just slightly more regularity holds.
\begin{theorem}\label{thm:fdidm}
Fix any $\lambda_n=o(n)$.
Suppose that in addition to the assumptions of \cref{thm:iidm}, in a neighborhood of $\theta_0$, $\psi(\theta)$ is thrice continuously differentiable and $f(Z;\theta)$ is almost surely thrice continuously differentiable in $\theta$ with a third-order derivative that is bounded in operator norm by an integrable function of $Z$.
Then $$n\hat V_n^\FDIDM\to_pV_0.$$
\end{theorem}
It may seem surprising that a constant $\lambda_n$ suffices or that $\lambda_n$ is even allowed to grow, but that can be seen as an artifact of the fact we did not normalize the sum over the data in \cref{eq:regmle} by $1/n$. If we did normalize, it would be equivalent to rescaling $\lambda$ by $n$, so that $o(n)$ becomes $o(1)$, \ie, requiring a vanishing increment for the finite differencing. Nonetheless, writing \cref{eq:regmle} as we did is very convenient, as it matches how one usually applies optimization algorithms such as stochastic gradient descent to training objectives, and it makes the choice of $\lambda_n$ for \cref{eq:fdidm} very easy: just fix some constant and do not worry about the scaling with $n$. For example, setting $\lambda_n=1$ suggests a very simple-looking 95\%-confidence interval: $\bracks{\hat\psi_n\pm1.96\sqrt{\hat\psi_n(1)-\hat\psi_n}}$. Note that it is not necessarily \emph{better} to choose smaller $\lambda$: the smaller $\lambda$ the closer $\hat V_n^\FDIDM$ is to $\hat V_n^\IIDM$, but that need not mean it is a better estimate (see numerical illustration in \reffig{convergence}). Finally, note that \cref{eq:fdidm} is but one way to make a finite-difference approximation of a derivative, and other finite-difference formulae for derivatives (see ch. 4 of \citealp{burden2015numerical}) such as central differences could possibly be used.

\begin{remark}[Regression Using Squared Error Loss]
When training regression models we usually minimize over model parameters (\eg, neural net weights) the sum over the data of squared error loss, $\ell((x,y);\theta)=(y-g_\theta(x))^2$. This differs from the corresponding Gaussian log likelihood by a factor of $-\frac{1}{2\sigma^2}$ (and some constants that do not matter), where $\sigma^2$ is the residual variance of $Y$ given $X$. Therefore, to apply IDM, all we should do is simply regularize the sum-of-squared-errors \emph{minimization} problem by $-2\sigma^2\lambda\psi(\theta)$, as that would be equivalent to dividing the log likelihood part by $-2\sigma^2$. Of course, we do not know $\sigma^2$, but we can estimate it by $\hat\sigma_n^2=\frac1n\sum_{i=1}^n(y-g_{\hat\theta_n}(x))^2$, that is, the minimum average sum of squared errors. Since $\hat\sigma_n^2\to_p\sigma^2$, as it is in fact the MLE estimate for $\sigma^2$, the asymptotic guarantees of \cref{thm:iidm,thm:fdidm} will continue to hold after this rescaling. Note that the standard errors given correspond to the MLE formulation of least-squares (usual standard errors) rather than the $M$-estimation formulation thereof (so-called robust or sandwich standard errors).
\end{remark}

\begin{remark}[Using IDM to Compute the Fisher Information]
A by-product of the proof of \cref{thm:iidm} is that, if we looked at the (vector-valued) derivative $\hat W_n=\frac{\partial}{\partial\lambda}\hat\theta_n(\lambda;\psi)\bigr\vert_{\lambda=0}=\lim_{\lambda\to0}\frac1\lambda\prns{\hat\theta_n(\lambda;\psi)-\hat\theta_n}$, then $n\hat W_n 
\to_p I(\theta_0)^{-1}\nabla\psi(\theta_0)$. Therefore, if we set $\psi(\theta)=\theta_i$, \ie, the $i\thh$ component of $\theta$, then $\hat W_n$ converges to the $i\thh$ column of $I(\theta_0)^{-1}$. Thus, by regularizing each component of $\theta$ in turn, we obtain the whole matrix.

Nonetheless, the whole raison d'\^etre of IDM is to avoid working directly with the parameter vector $\theta$ altogether, and simply propagate its uncertainty automatically via the MLE optimization problem. For example, if we consider neural net regression, IDM would never make explicit reference to the vector of weights itself, only to the trained prediction model and its prediction performance on data. The above, wherein we compute the uncertainty in $\theta$ directly, stands in contradiction to this. Nonetheless, it can be a useful observation when inference on $\theta$ itself is for some reason of interest.
\end{remark}

\subsection{Extension to Multivariate Evaluations}\label{sec:multivar}

We have so far focused on scalar evaluations for ease of presentation and as it covers the most important cases. We now show how our method easily extends to the multivariate case, where $\psi(\theta)=(\psi\s1(\theta),\dots,\psi\s K(\theta))\in\RR^K$. The reason it may not suffice to run IDM separately for each component is that we may be interested in the \emph{covariance} of the evaluations. Under the appropriate conditions, the extension of the delta method for MLE (\cref{eq:deltamethod}) to multivariate evaluations is
\begin{equation}\label{eq:deltamethod2}
\sqrt{n}(\hat\psi_n-\psi_0)\rightsquigarrow\Ncal(0,\,V_0),\quad
V_0=J(\theta_0)\tr I^{-1}(\theta_0)J(\theta_0),
\end{equation}
where $J_{ij}(\theta)=\frac{\partial}{\partial\theta_i}\psi\s j(\theta)$ is the $K\times d$ Jacobian of $\psi(\theta)$.

Our extensions of IIDM and FDIDM to multivariate evaluations are as follows:
$$
\Delta_{ij}(\lambda)=\frac1\lambda\prns{\psi\s i(\hat\theta_n(\lambda;\psi\s j))-\psi\s i(\hat\theta_n)},~~
(\hat V_n^\IIDM)_{ij}=\lim_{\lambda\to0}\Delta_{ij}(\lambda),~~
(\hat V_n^\FDIDM)_{ij}=\Delta_{ij}(\lambda_n).
$$
\begin{theorem}\label{thm:multivar}
$n\hat V_n^\IIDM\to_pV_0$ under the conditions of \cref{thm:iidm}, and
$n\hat V_n^\FDIDM\to_pV_0$ under the conditions of \cref{thm:fdidm}, both as $K\times K$ matrices.
\end{theorem}

Surprisingly, this shows one need only solve $K+1$ (possibly) regularized MLEs to get the full $K\times K$ covariance. (See Alg.~\ref{alg:fdidm2} in supplement.)

\subsection{Handling Non-differentiable Evaluations and Evaluation Uncertainty}

So far we have assumed that the evaluation function is a known and differentiable function. Both statements may be false when we are interested in evaluating average performance on a population but we only have a finite evaluation data set and unit performance is not differentiable. 

Specifically, let $W_1,\dots,W_m\sim W$ denote the evaluation data set (which may be the same as the training set or otherwise dependent or it may be an independent data set) and let $h(w;\theta)$ the unit evaluation function. 
Consider the empirical evaluation map
$$
\ts\psi(\theta)=\frac1m\sum_{j=1}^mh(W_i;\theta).
$$
If $h(W;\theta)$ is almost surely not differentiable in $\theta$, then $\psi$ is also almost surely not differentiable, which poses a challenge. We will show, however, that even though $\psi$ is not differentiable (which would break the usual delta method), FDIDM actually remains valid, \emph{without any changes to the method}, provided certain on-average-differentiability holds.

To motivate the challenge of nondifferentiability and the plausibility of on-average-differentiability,
consider an example where $g_\theta(x)$ represents an order quantity to stock in context $x$ and $w=(x,d)$ represents features and demand. If $h((x,d);\theta)=\max\{d-g_\theta(x),0\}$ then $\psi(\theta)$ quantifies average unmet demand, but $h$ is not differentiable. 
Other non-differentiable examples include evaluating regression and classification models' performance using non-differentiable utility functions. 
While $h$ may not be differentiable and hence neither $\psi$, it may still be plausible that its expectation $\E[h(W;\theta)]=\E[\psi(\theta)]$ is differentiable. For example, if the distribution of demand conditioned on features is continuous, then the derivative of $\E[h(W;\theta)]$ in the example of average unmet demand will be the average of the conditional cumulative distribution function at $g_\theta(x)$ times $-\nabla_\theta g_\theta(x)$, and the second derivative will involve the conditional density.

We next show FDIDM \emph{still} works with non-differentiable $\psi$, given some on-average-differentiability.
\begin{theorem}\label{thm:fdidm2}
Consider $m=\Omega(n)$. Fix $\lambda_n=\lambda>0$ constant.
Suppose the assumptions of \cref{thm:fdidm} hold, that \cref{eq:mleconv} holds, that $h(W;\theta)$ is almost surely $L$-Lipschitz in $\theta$, and that for some $M>0$,
$$
\lim_{\epsilon\to0}\Prb{\text{On $\{\theta:\magd{\theta-\theta_0}\leq\epsilon\}$, $h(W;\theta)$ is twice differentiable in $\theta$ with $\magd{\nabla_\theta^2h(W;\theta)}\leq M$}}=1.
$$
Then
$$
\prns*{\hat V_n^\FDIDM}^{-1/2}\prns*{\hat\psi_n-\psi_0}\rightsquigarrow\Ncal(0,1).
$$
\end{theorem}

In the above example of average unmet demand, $L$ and $M$ would be bounds of the gradient and Hessian of $g_\theta(x)$ in $\theta$, and a sufficient condition for the assumption to hold would be that $g_\theta(x)$ is boundedly differentiable in $x$ for $\theta$ in a neighborhood of $\theta_0$ and $W=(X,D)$ has a continuous distribution.

Although the asymptotic variance of $\hat\psi_n-\psi_0$ is now different (in particular $\hat V^{\mathrm{DeltaMethod}}_n$ in \cref{eq:Vn} may be ill-defined), \cref{thm:fdidm2} shows that $\hat V_n^\FDIDM$ actually remains consistent for this new asymptotic variance. Thus, it provides a consistent estimate of standard errors and it still gives calibrated confidence intervals (note $\psi_0$ is now \emph{random} but we can still have a confidence interval for it).

In some cases we may want to directly conduct inference on the population version of the evaluation, $\psi^*_0=\E[h(W;\theta_0)]$.
To do this, all we have to do is simply also add the uncertainty due to finite evaluation data set.
Under standard regularity conditions, we have
\begin{align*}
&\ts\prns*{\hat V_m^\psi}^{-1/2}\prns*{\psi_0-\psi^*_0}\rightsquigarrow\Ncal(0,1),
\quad\text{where}\quad\hat V_m^\psi=\frac1{(m-1)m}\sum_{j=1}^m\prns*{h(w;\hat\theta_n)-\psi(\hat\theta_n)}^2.
\end{align*}
Therefore, provided the training and evaluation data sets are independent,
\begin{align*}
&\prns*{\hat V_n^\FDIDM+\hat V_m^\psi}^{-1/2}\prns*{\psi(\hat\theta_n)-\psi^*_0}\rightsquigarrow\Ncal(0,1).
\end{align*}
If not independent, then $\sqrt{\hat V_n^\FDIDM}+\sqrt{\hat V_m^\psi}$ provides a consistent upper bound on the standard error.

\subsection{Implementation}

FDIDM is given in pseudocode in Alg.~\ref{alg:fdidm}. 
Given an objective function $\mathcal{L} := \sum_i^n \log f(Z_i; \theta)$, 
evaluation function $\psi$, and scalar width $\lambda$, 
FDIDM returns the estimated variance of $\psi(\hat{\theta}_n)$. 
The first step is to maximize the original objective w.r.t. $\theta$.  
Usually, this task has already been solved as this is the trained predictive model. 
Then, maximize the $\psi$-regularized objective w.r.t. $\theta$. 
Finally, return the estimated variance using the finite-difference method 
evaluated at $\lambda=0$. 
See Appendix~\ref{sec:appendix_mv_psi} for the corresponding algorithm when $\psi$ is multivariate. 

In practice, one can further reduce the computational cost of 
FDIDM due to the fact that the $\psi$-regularized objective 
can be made arbitrarily close to original objective by choosing $\lambda$ small enough, 
subject to numerical instability at extremely small values. 
Specifically, when using stochastic gradient ascent in FDIDM, 
once the optimum $\hat{\theta}_n$ has been found, 
only a small number of gradient updates may be required to also find $\hat{\theta}_n(\lambda)$. 

FDIDM also admits non-gradient-based approaches. 
Consider the case that $\psi$ is a simulator that 
takes a fitted model and returns 
a set of evaluations and no gradient. 
Then the $\psi$-regularized objective may optimized by gradient-free methods such as Nelder-Mead \cite{nelder1965simplex}
and Bayesian optimization \cite{frazier2018bayesian}.

\begin{algorithm}[t!]
\SetAlgoLined
\KwIn{Learning objective $\objective$, evaluation $\psi$, scalar $\lambda$} 
\DontPrintSemicolon
  \SetKwFunction{FMain}{FDIDM}
  \SetKwProg{Fn}{Function}{:}{}
  \Fn{\FMain{$\mathcal{L}$, $\psi$, $\lambda$}}{
        $\hat{\theta}_n \gets \arg \max_\theta \objective(\theta)$ \tcp*{optimize learning objective}
        $\hat{\theta}_n(\lambda) \gets \arg \max_\theta \objective(\theta) + \lambda \psi(\theta)$ \tcp*{optimize $\psi$-regularized objective}
        \KwRet $\frac{1}{\lambda}(\psi(\hat{\theta}_n(\lambda)) - \psi(\hat{\theta}_n))$ \tcp*{estimated variance of $\psi(\hat{\theta}_n)$}
  }
 \caption{Finite-difference implicit delta method (FDIDM)}
 \labelalg{fdidm}
\end{algorithm}

\section{Alternatives for Uncertainty Quantification and Related Work}
\labelsec{related}

Uncertainty quantification in machine learning is a topic of major interest 
due to the need to make downstream inferences and decisions 
based on the predictions of large-scale networks trained on massive datasets 
in either a frequentist~\cite{osband2016deep, giordano2019swiss, nilsen2022epistemic} or Bayesian fashion~\cite{blundell2015weight, gal2016dropout, daxberger2021laplace}. 
Our focus here is on methods 
that can flexibly isolate epistemic uncertainty 
in an evaluation, 
representing data sampling uncertainty of that evaluation under a given model. 
In cases where the total uncertainty for predictions is desired, 
a broader set of methods may be brought to bear, 
such as conformal prediction~\cite{vovk2005algorithmic, romano2020classification, angelopoulos2021uncertainty}, Platt scaling~\cite{platt1999probabilistic, guo2017calibration}, 
or indeed, any of the aforementioned statistical methods 
used in conjunction with a term or terms for aleatoric uncertainty. 
An exhaustive account of the literature is outside the scope of this paper. 
We highlight the principal ideas and points of contact with our work.

\paragraph{The Bootstrap} 

The bootstrap simulates sampling from the true data generating distribution by resampling from the observed dataset (see, \eg, \cite{efron2021computer} for an introduction and \cite{kosorok2008introduction} for theory on when it works). 
The key advantage is that it enables 
general-purpose and easy-to-implement 
uncertainty quantification for estimators.
It comes at a high computational burden 
because the estimator, which may comprise a model-fitting algorithm and prediction, 
needs to be executed many times. 
In the context of deep learning, 
many useful adaptations of the bootstrap and the related jackknife 
have been proposed to increase its computational efficiency \cite{osband2015bootstrapped, osband2016deep, giordano2019swiss}. 
Maintaining an ensemble of models as a representation of the variability 
of the evaluation is an appealing intuition 
that does not restrict one to local approximations, 
and may be combined with local approximations where necessary.

\paragraph{The Functional Delta Method} 
The delta method \cite{doob1935limiting} is a classic approach that is widely used 
with small models with an analytic Fisher information matrix 
(\eg, linear regression) 
and, more recently, auto-differentiation unlocks 
the delta method for a larger class of models \cite{nilsen2022epistemic}. 
The bottleneck is the need to calculate then 
invert the Fisher information matrix, 
for which there are various approximations~\cite{pearlmutter1994fast, lorraine2020optimizing}. 
The delta method applies to a wide range of (differentiable) estimators subject to 
regularity conditions that ensure asymptotic normality of the parameter estimates 
and this constraint carries over to the implicit delta method. 
The functional delta method extends the delta method to evaluations of infinite-dimensional parameters (see Ch.~12 \citep{kosorok2008introduction}) but is usually restricted to analytically deriving influence functions in theory by differentiating the population estimand with respect to distributions and then approximating the influence function by plugging in estimates of unknown nuisances \citep{ichimura2022influence,chernozhukov2018double}.

\paragraph{Bayesian Uncertainty Quantification} 

Tractable methods for approximate Bayesian inference in neural networks,  
such as variational inference in feed-forward nets \cite{blundell2015weight}, autoencoders \cite{kingma2013auto, rezende2014stochastic}, normalizing flows \cite{rezende2015variational}, dropout uncertainty \cite{gal2016dropout}, 
stochastic gradient Langevin dynamics \cite{welling2011bayesian} and related approaches,  
present an impressive range of options for uncertainty quantification. 
In cases where it is sufficient to 
only consider a single mode in the posterior, 
local methods can prove useful. 
In particular, an alternative interpretation of the delta method 
is as a special case of the Laplace approximation 
to Bayesian inference, 
where \cref{eq:Vn} arises 
in the posterior predictive distribution 
for a local multivariate Gaussian approximation around the maximum \emph{a~posteriori} estimate. 
Several recent works 
have investigated the potential 
of the Laplace approximation 
as a way to avoid having to characterizing the full posterior 
in deep networks \cite{daxberger2021laplace, khan2019approximate, immer2021improving}. 
IDM can provide another way 
to perform a Laplace approximation 
and may be orthogonally combined with the above methods.

\section{Experiments}
\labelsec{experiments}

In this section, we evaluate finite-difference implicit delta method (FDIDM) 
on a range of tasks that require confidence intervals.\footnote{The source code is available at \url{https://github.com/jamesmcinerney/implicit-delta}.} 
Our goal is to quantify the extent to which FDIDM 
applies in practice and how it compares to 
alternative methods. 
We start with 1D synthetic data in Sec.~\ref{sec:exp_1d} 
where we apply a neural net to recover known functions 
from small datasets. 
Then, in Sec.~\ref{sec:exp_utility}, 
we consider the task of inferring average utility 
under a neural net trained on a set of real-world benchmark datasets. 
In Sec.~\ref{sec:exp_vae}, 
we apply FDIDM to variational autoencoders 
and use the implicit delta perspective to understand the effect of KL down-weighting. 
We find that the motivation and convergence properties 
of FDIDM 
are empirically observed and this may be useful to practitioners 
seeking to quantify the epistemic uncertainty of complex models on a variety of regression and classification tasks.

\begin{figure*}[t!]%
\centering%
\begin{subfigure}[b]{0.45\textwidth}%
\includegraphics[height=3.5cm]{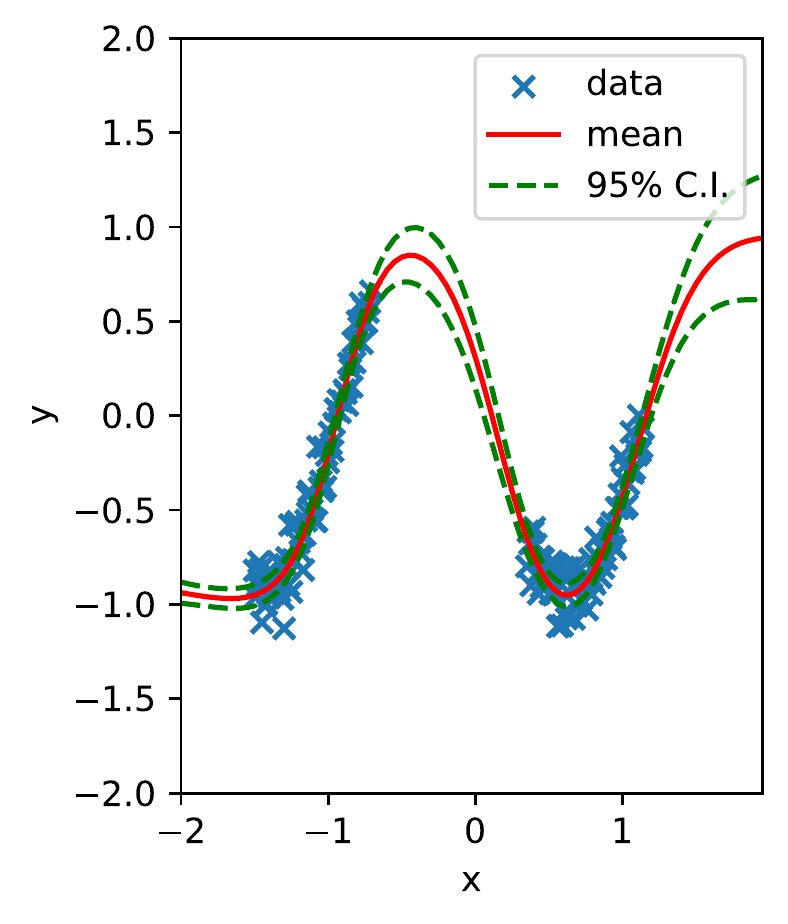}\includegraphics[height=3.5cm]{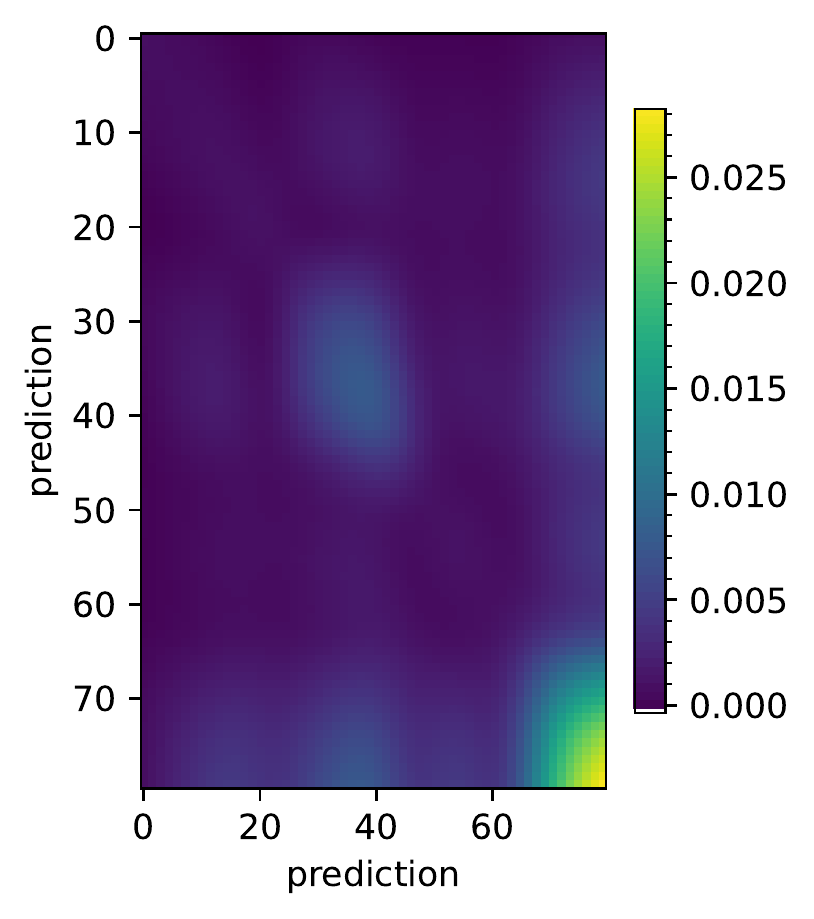}%
\vspace{-2ex}\caption{\capsize \namep (this paper)}%
\end{subfigure}%
\begin{subfigure}[b]{0.45\textwidth}%
\includegraphics[height=3.5cm]{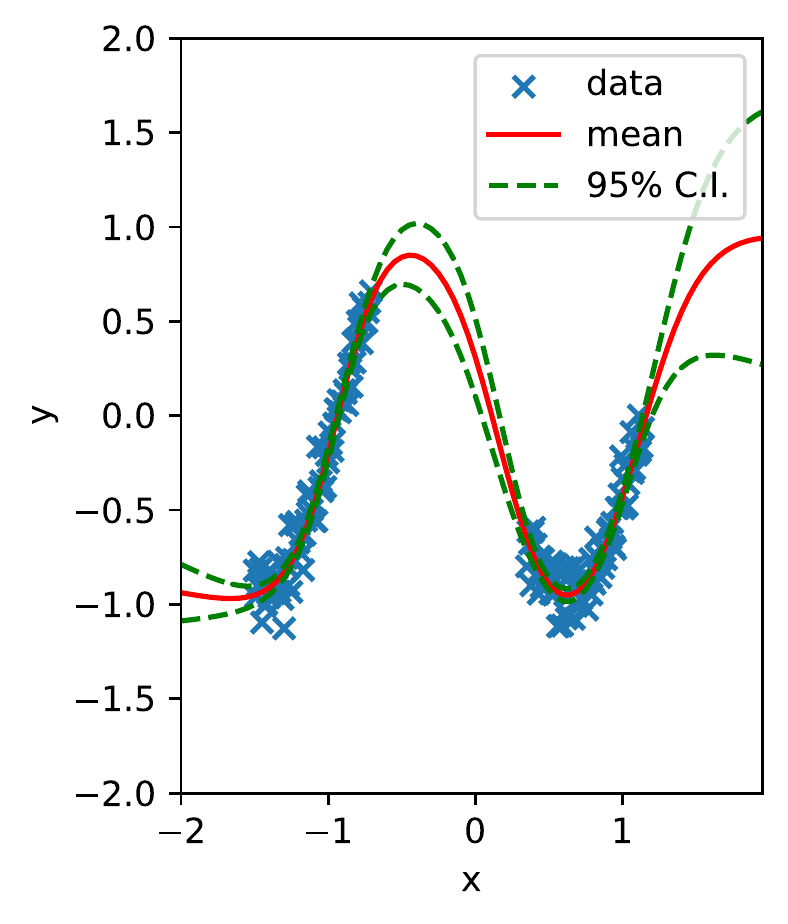}\includegraphics[height=3.5cm]{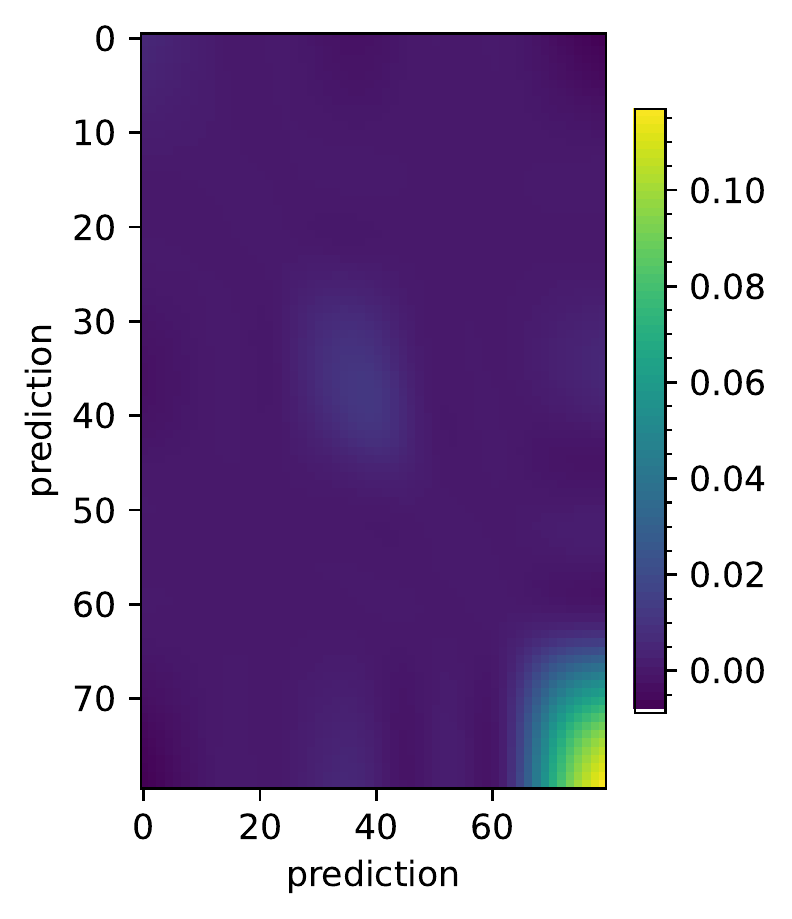}%
\vspace{-2ex}\caption{\capsize Delta Method}%
\end{subfigure}\\[1ex]%
\begin{subfigure}[b]{0.45\textwidth}%
\includegraphics[height=3.5cm]{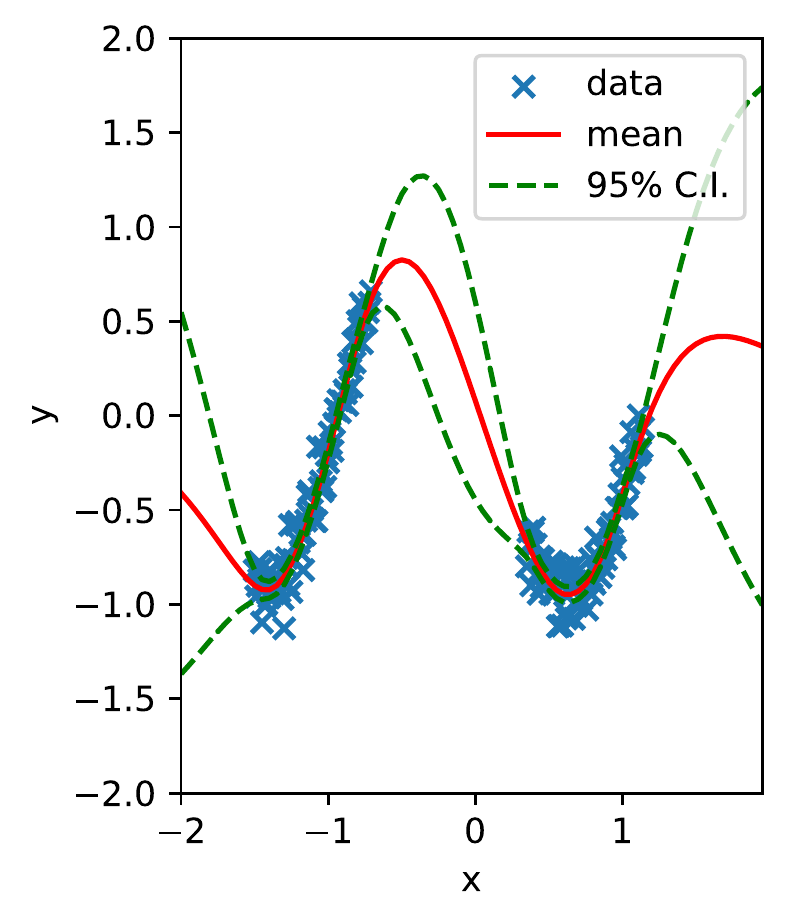}\includegraphics[height=3.5cm]{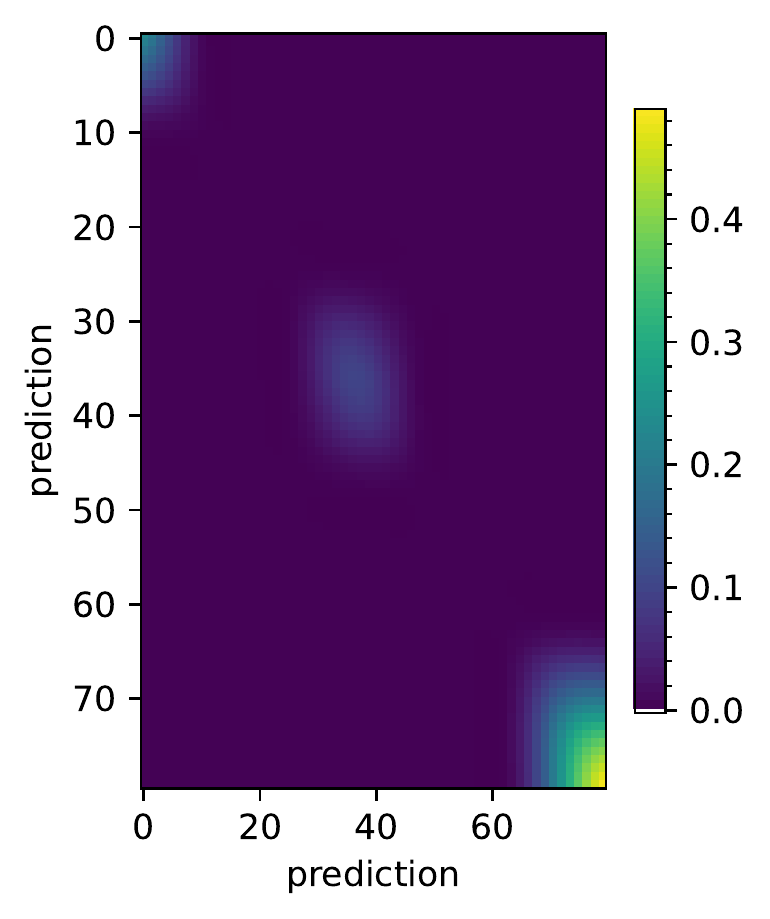}%
\vspace{-2ex}\caption{\capsize GP-Matern52}%
\end{subfigure}%
\begin{subfigure}[b]{0.45\textwidth}%
\includegraphics[height=3.5cm]{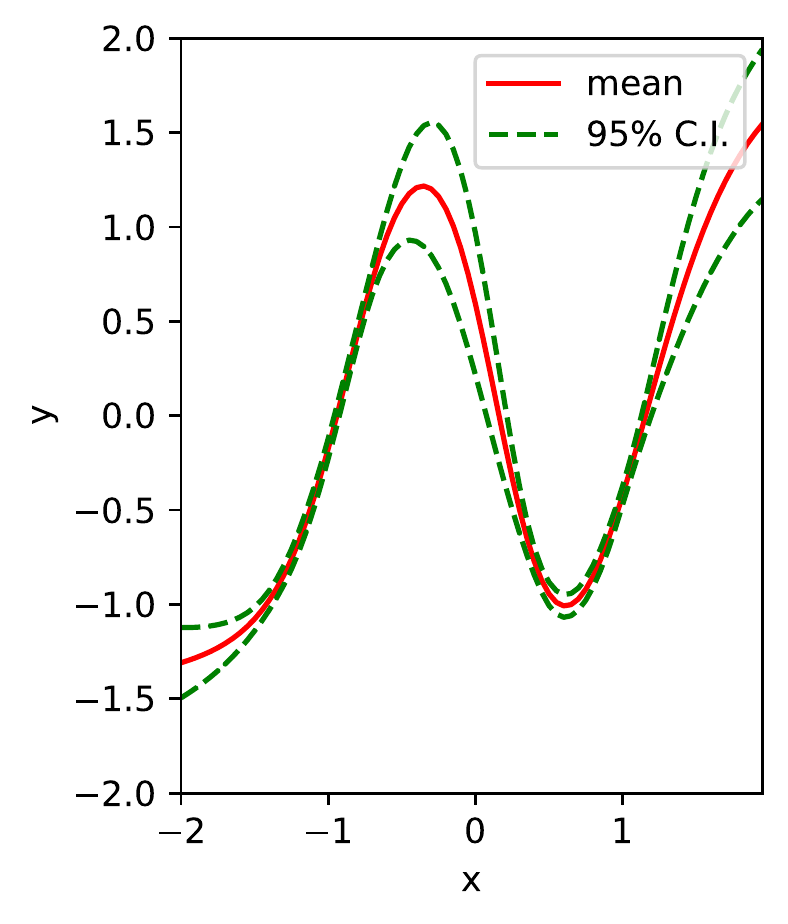}\includegraphics[height=3.5cm]{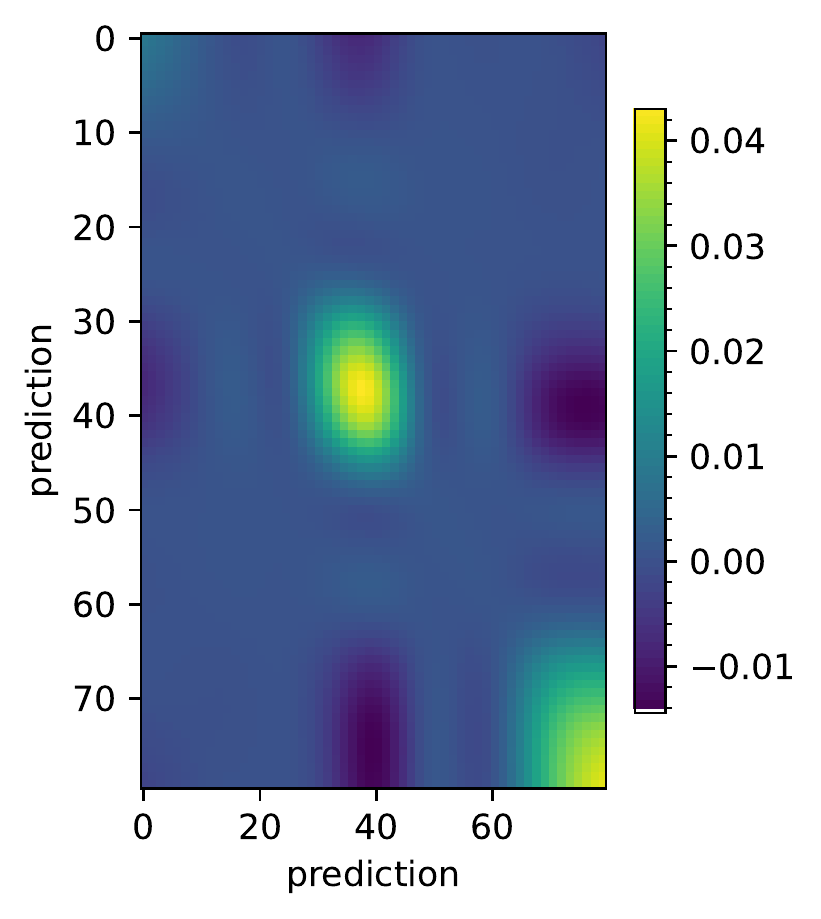}%
\vspace{-2ex}\caption{\capsize True sampling uncertainty by simulation}%
\end{subfigure}%
\caption{Fits along with uncertainty bounds and estimated prediction-covariance matrix for data generated from $y = -\sin(3x - \frac{3}{10}) + \frac{1}{10}\epsilon$, where $\epsilon \sim \mathcal{N}(0,1)$}%
\labelfig{sin_fit}%
\end{figure*}

\begin{figure*}[t!]%
\centering%
\hfill\begin{subfigure}[b]{0.43\textwidth}%
\includegraphics[width=1\linewidth]{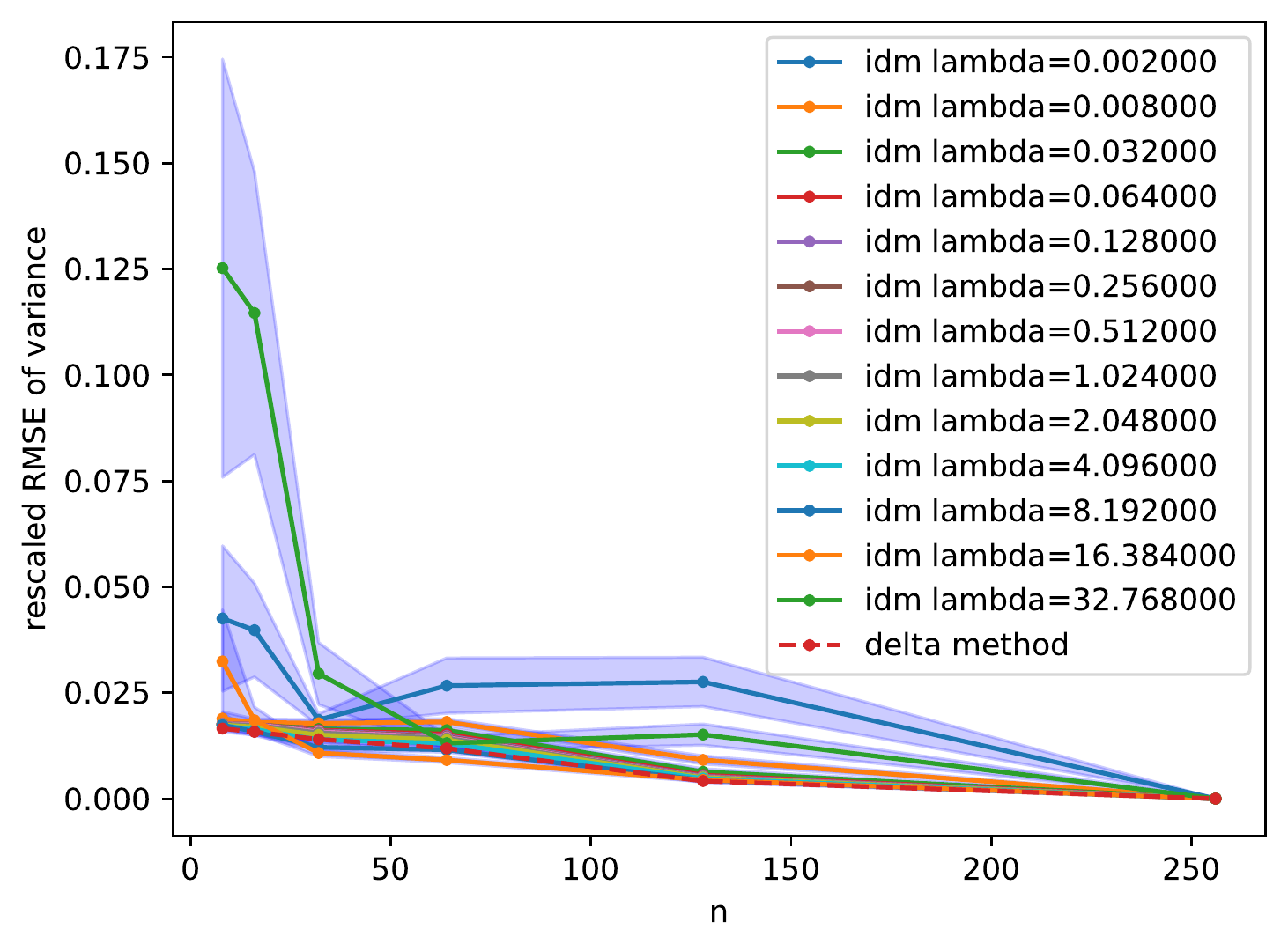}
\caption{\capsize Number of data points as independent variable.}%
\end{subfigure}\hfill%
\begin{subfigure}[b]{0.43\textwidth}%
\includegraphics[width=1\linewidth]{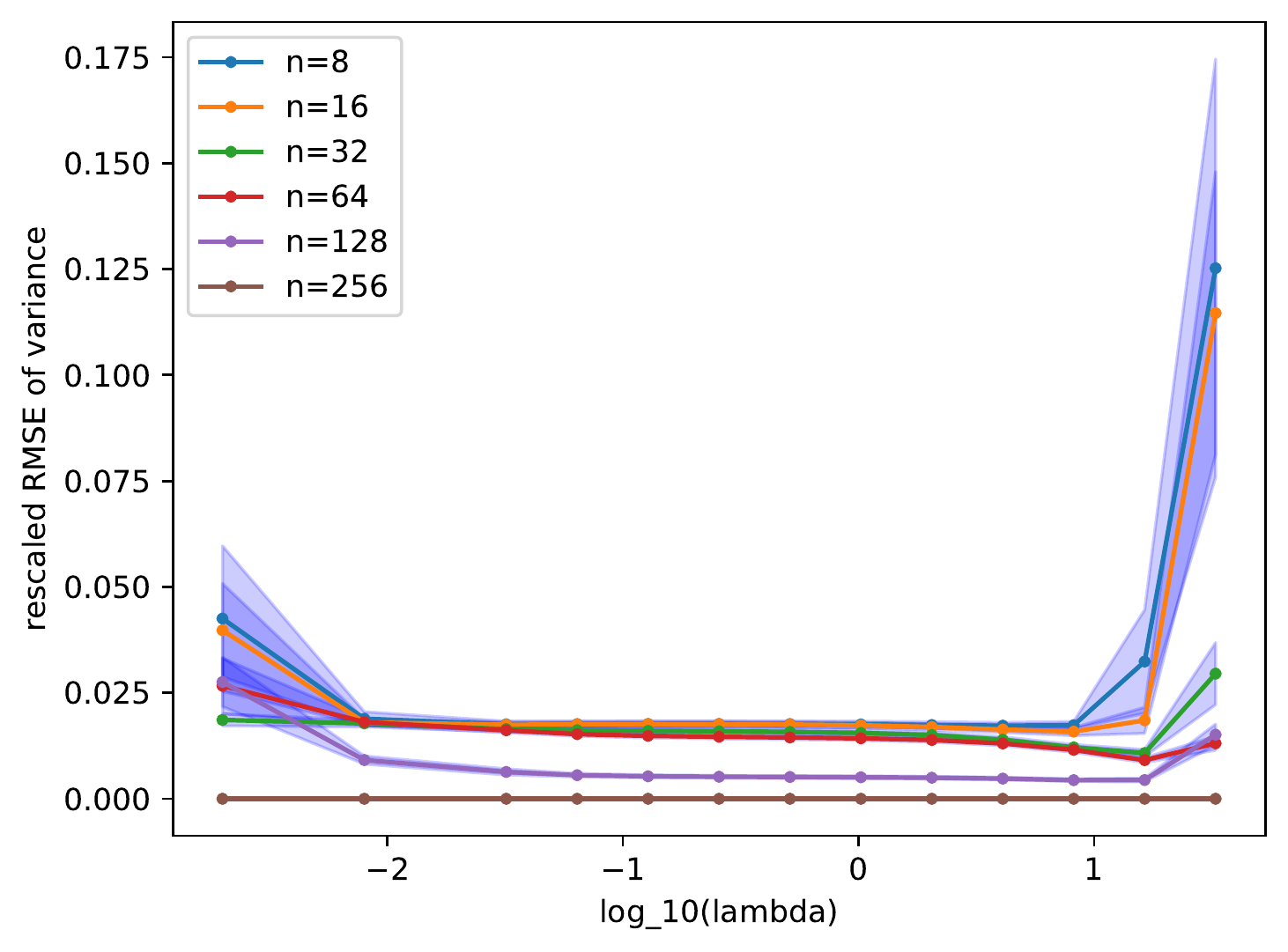}
\caption{\capsize $\lambda$ as independent variable.}%
\end{subfigure} 
\caption{Convergence of IDM in the quadratic task for different values of $n$ and $\lambda$. There is a wide dynamic range of acceptable $\lambda$.}\hfill%
\labelfig{convergence}%
\end{figure*}

\subsection{1D Synthetic Examples}
\label{sec:exp_1d}

We consider known quadratic and sinusoidal functions 
from which we draw a random dataset. 
Fig.~\ref{fig:sin_fit} 
gives the data generating stochastic function for a $\sin$ wave 
and the resulting fits for FDIDM, the classic delta method, a Gaussian process (GP) with Matern-52 kernel, as well as 
simulation from the true data generating function. 
(Appendices~\ref{sec:further_exp}~and~\ref{sec:appendix_exp} provide the results on the quadratic function and further experimental details, respectively.) 
The quadratic example has evenly dispersed input data 
and there is close alignment between the methods. 
The $\sin$ wave is more challenging because it requires extrapolation -- also known as ``in-between'' uncertainty in \cite{foong2019between} -- 
from outside the ranges of given inputs. 
Results for IDM, DM, and simulation are all based on estimates using a neural net with 1~hidden layer of 50 $\texttt{tanh}$ units.\footnote{This architecture is in line with \cite{foong2019between}, which also provided the basis for our $\sin$ example.} 
It should be noted that the GP is not trying to estimate the frequentist sampling variance (shown in the simulation results) but rather the Bayesian posterior uncertainty (although they can coincide asymptotically; \citealp[theorem 11.5]{wasserman}); we include it largely for a qualitative comparison to a popular epistemic-uncertainty quantification method.
In particular, unlike the IDM, DM, and simulation results, the GP does not yield an interval around the neural-net based mean estimate and instead has a different mean function\footnote{We also applied the $\arccos$ kernel in a GP which imitates a neural network~\cite{cho2009kernel} but found that the Matern-52 kernel inferred a mean that was closer in practice to the mean inferred by the neural net.} 

As expected, IDM agrees most with the delta method 
while the GP overestimates uncertainty, particularly for extrapolation at the outer edges. 
The corresponding full covariance matrix of the 
predictions is also given in Fig.~\ref{fig:sin_fit}. 
All the methods recover the high-level structure of covariance for both examples, 
though the scale factors differ considerably. 

\reffig{convergence} shows the convergence of the root mean squared error of IDM w.r.t. the true variance as determined by 50 resamples from the data generating distribution in the quadratic task. The squared errors are rescaled by $n^2$ to account for decreasing scale ($\frac{1}{n}$) of the true the variance as $n$ grows. Shaded error bars indicate one standard error. Convergence for the standard delta method is also shown for reference. We find there is a wide dynamic range of acceptable values of $\lambda$. Small values of $\lambda<0.01$ perform poorly, likely due to numerical instability, but performance improves for larger $\lambda$. The setting $\lambda=0.512$ even outperforms the delta method. These findings support the implication of Theorem~\ref{thm:fdidm} indicating that convergence holds as long as $\lambda$ grows sublinearly to $n$.

\subsection{Confidence in Predicted Cost Downstream of Classification}
\label{sec:exp_utility}

A set of classification tasks are fitted with a neural net with one hidden layer and 50 $\texttt{tanh}$ hidden units. 
In this setting, we wish to calculate confidence intervals over total cost 
in a downstream task under predictions from the network. 
An arbitrary cost function is set up, in this case, 
the average cross entropy of the observations on a held-out validation dataset, 
though in practice we could have a wide variety of cost functions 
relating to the task downstream of the classifier. 
It is challenging to form a confidence interval for even this simple cost function 
because it is a function of predictions from the network. 
Under this scenario, it is typical to make a bootstrapping estimate, requiring $B$ times the cost of training the network (here, we use $B=50$). 
FDIDM is also applicable in this setting. We show both methods on MNIST image classification~\cite{lecun1998gradient} and a set of UCI benchmark datasets~\cite{dua2019uci} in Fig.~\ref{fig:utility}. 
We find that FDIDM has good coverage for a fraction of the computational cost of the bootstrap estimate. 
Specifically, a time complexity comparison is provided in Table~\ref{tab:time_complexity}.\footnote{Run time was measured on a MacBook Pro 2.3 GHz Quad-Core Intel Core i7 with 32 GB RAM.}

\begin{table}[t!]\vspace{-1.7em}
  \caption{Run time (seconds)}
  \label{tab:time_complexity}
  \centering
  \begin{tabular}{lllll}
    \toprule
     & Vehicle     & Waveform & Satellite & MNIST \\
    \midrule
    IDM & 39 & 129 & 111 & 303 \\
    Bootstrap     &806 & 2,334 & 3,192 & 7,164      \\
    \bottomrule
  \end{tabular}
\end{table}

\subsection{Down-Weighting KL in Variational Autoencoders}
\label{sec:exp_vae}

\begin{figure*}[t!]%
\centering%
\begin{minipage}[b]{.48\textwidth}
  \centering
  \vspace{11pt}
  \includegraphics[width=.8\linewidth]{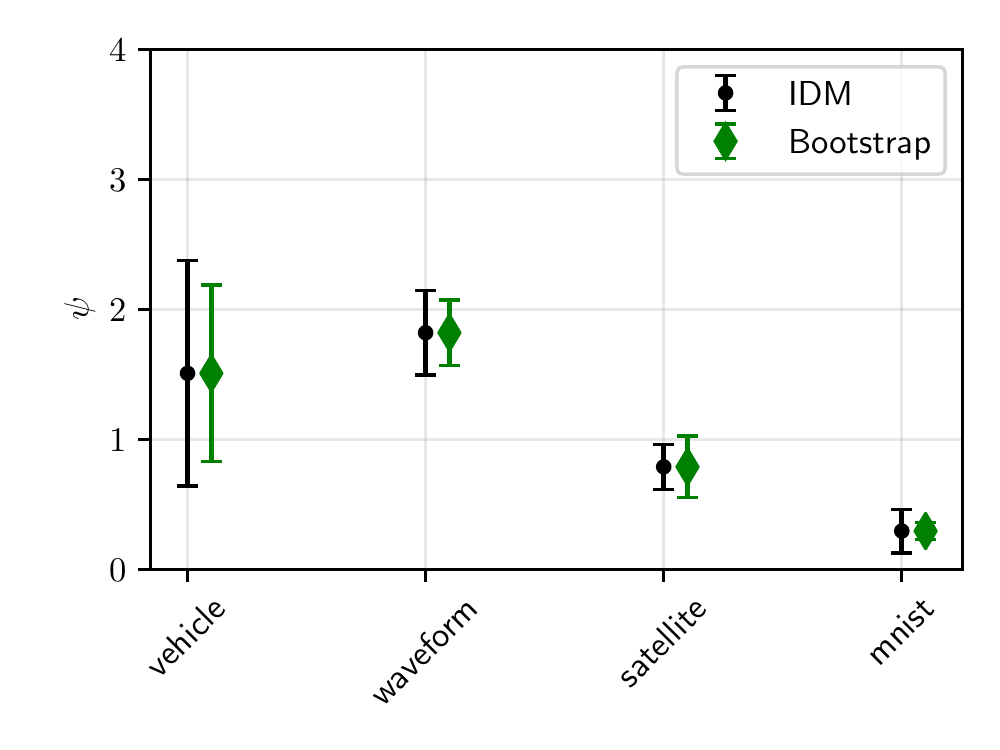}\vspace{-6pt}
  \caption{Predicted distribution of utility in classification on benchmark datasets. Datasets are shown in order of number of examples.}
  \labelfig{utility}
\end{minipage}%
\hfill%
\begin{minipage}[b]{.50\textwidth}
  \centering
  \vspace{-8pt}
  \includegraphics[width=.8\linewidth]{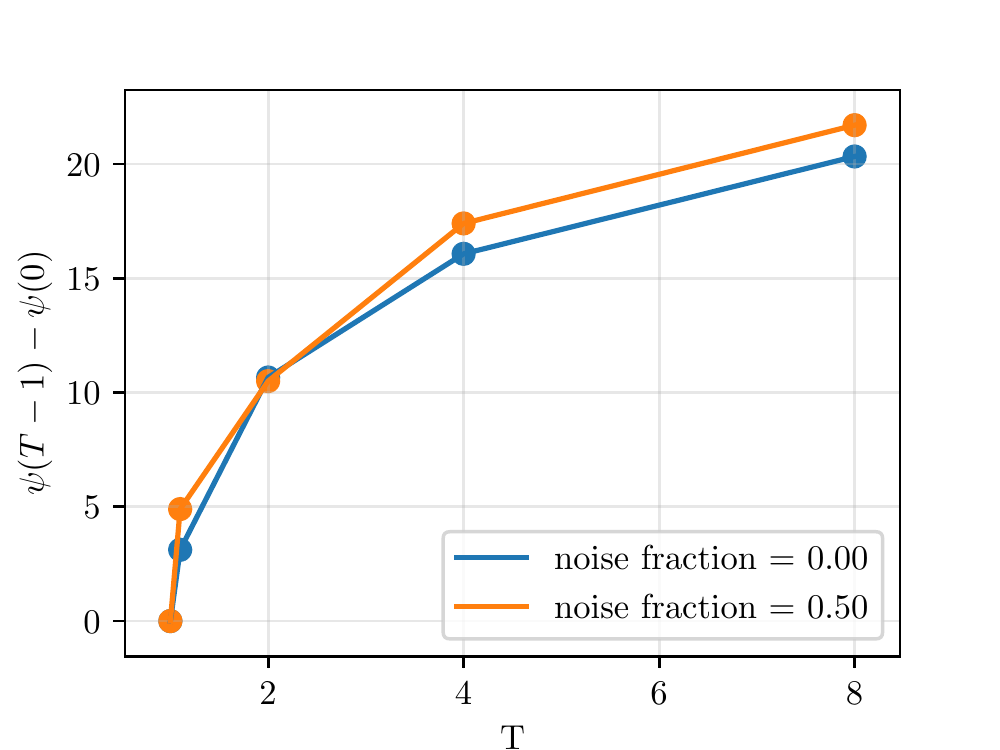}
  \vspace{-1pt}
  \caption{Improvement in reconstruction quality over the unweighted likelihood as a function of $T$.\\}
  \label{fig:vae_idm_T}
\end{minipage}%
\end{figure*}

The variational autoencoder (VAE) is a prominent example of 
approximate inference in deep generative models \cite{kingma2013auto, rezende2014stochastic}. 
In practice, it has been observed that down-weighting the KL term in the variational objective by a factor $\frac{1}{T}$, where $T > 1$, 
results in significantly better accuracy on held-out test data \citep{zhang2018noisy, ashukha2020pitfalls}. 
This is closely related to other ways of reweighting the prior and likelihood terms in approximate inference 
such as data augmentation \cite{osawa2019practical} and the cold posterior effect \cite{wenzel2020good}. 
Various explanations for the benefit of setting $T > 1$ have been posited, 
\eg, model misspecification. 
Here, we briefly explore an IDM interpretation of this phenomenon. 

Observing that any objective does not change its critical points under (non-zero) rescaling, 
it holds that down-weighting the KL term yields the same 
optimization problem as up-weighting the reconstruction error by $T$. 
We consider this setting and define $\psi$ to be the reconstruction quality 
(\ie, negative reconstruction error) while 
optimizing the standard (unweighted) evidence lower bound defined by the VAE. 
Rearranging the terms in \cref{eq:fdidm}, letting $\lambda_n := T-1$, 
we have 
\begin{flalign}
	\hat\psi_n(\lambda_n) &= (T-1) \hat V_n^\FDIDM + \hat\psi_n(0). \label{eq:vae_idm}
\end{flalign}

The two immediate implications of \cref{eq:vae_idm} for fixed $T>1$ of $o(n)$ are that, 
\begin{enumerate}[itemsep=0mm,leftmargin=*,topsep=0mm]
	\item the reconstruction quality for the objective implied by variational inference  
	is upper bounded by that of the objective with up-weighted likelihood; and,
	\item the higher the variance in the reconstruction quality, as determined by the dataset and model, 
	the greater the benefit of up-weighting the likelihood term in variational inference. 
\end{enumerate}
Since (1) is already supported by existing empirical work, 
we focus on evaluating whether~(2) also holds in practice. 
To do this, we artificially increase the variance of the reconstruction quality 
by perturbing a proportion of the dataset 
and compare $\hat{\psi}_n$ for various $T$. 
Fig.~\ref{fig:vae_idm_T} shows the results for applying a VAE 
to the MNIST dataset 
after adding i.i.d. noise $\epsilon \sim \mathrm{Uniform}(0,\frac{1}{20})$ 
to each pixel in a randomly chosen fraction of the images. 
The figure indicates that,  
as predicted by \cref{eq:vae_idm}, 
the gap in the reconstruction quality 
for $T > 1$ relative to $T=1$ increases 
as more variance is introduced. 
Higher values of $T$ do indeed result in better 
reconstruction quality and this advantage grows 
with the amount of variance. 
In sum, these findings 
are consistent with the hypothesis that 
the advantage from KL down-weighting 
may be explained as the residual 
between the $\psi$-regularized variational objective and 
the objective implied by the evidence lower bound, 
though further studies are required.

\section{Discussion, Limitation, and Conclusions}
\labelsec{discussion}

In this paper we develop the implicit delta method 
for forming calibrated confidence intervals 
via a careful regularization of the model objective. 
Like the delta method,
the method requires certain regularity conditions (Theorem~\ref{thm:iidm}) 
and for the MLE to be at a stable optimum, where perturbations around the optimum reliably capture sampling uncertainty. 
If this is not the case, 
\eg, the parameter 
has failed to converge or the 
objective itself is changing, 
it yields unreliable results. 
For this reason, IDM -- like the delta method and the bootstrap -- may be misleading for small data, and indeed uncertainty quantification with small data is fundamentally difficult. 
The most appealing feature of IDM is that 
it does not require the variance of the parameters 
to be made explicit, which also
suggests future research in exploring the 
compatibility of nonparametric models with IDM. 
There is also the potential to explore IDM in constrained MLE and in general $M$-estimation.

\section*{Acknowledgments}

We are grateful for the insightful comments 
of the anonymous reviewers and our colleagues at~Netflix.

\bibliographystyle{plainnat}
\bibliography{lit}

\section*{Checklist}

\begin{enumerate}

\item 
\begin{enumerate}
  \item Do the main claims made in the abstract and introduction accurately reflect the paper's contributions and scope?
    \answerYes{Proof of consistency in Sec.~\ref{sec:idm}; connections discussed in Sec.~\ref{sec:related}; evaluation in Sec.~\ref{sec:experiments}}
  \item Did you describe the limitations of your work?
    \answerYes{See Sections~\ref{sec:setup_problem} and~\ref{sec:discussion}}
  \item Did you discuss any potential negative societal impacts of your work?
    \answerNA{To the best of the authors' knowledge, our work does not have negative social impacts}
  \item Have you read the ethics review guidelines and ensured that your paper conforms to them?
    \answerYes{}
\end{enumerate}

\item 
\begin{enumerate}
  \item Did you state the full set of assumptions of all theoretical results?
    \answerYes{See Sec.~\ref{sec:idm} and Appendix~A}
        \item Did you include complete proofs of all theoretical results?
    \answerYes{See Sec.~\ref{sec:idm} and Appendix~A}
\end{enumerate}

\item 
\begin{enumerate}
  \item Did you include the code, data, and instructions needed to reproduce the main experimental results (either in the supplemental material or as a URL)?
    \answerYes{See online repository \url{https://github.com/jamesmcinerney/implicit-delta}}
  \item Did you specify all the training details (e.g., data splits, hyperparameters, how they were chosen)?
    \answerYes{See Sec.~\ref{sec:experiments}, Appendix~B~and~C, and source code}
        \item Did you report error bars (e.g., with respect to the random seed after running experiments multiple times)?
    \answerYes{}
        \item Did you include the total amount of compute and the type of resources used (e.g., type of GPUs, internal cluster, or cloud provider)?
    \answerYes{See Appendix~C}
\end{enumerate}

\item 
\begin{enumerate}
  \item If your work uses existing assets, did you cite the creators?
    \answerYes{}
  \item Did you mention the license of the assets?
    \answerNA{}
  \item Did you include any new assets either in the supplemental material or as a URL?
    \answerYes{Source code}
  \item Did you discuss whether and how consent was obtained from people whose data you're using/curating?
    \answerNA{}
  \item Did you discuss whether the data you are using/curating contains personally identifiable information or offensive content?
    \answerNA{}
\end{enumerate}

\end{enumerate}

\clearpage 

\appendix

\begin{center}\LARGE\bf
Supplementary Materials
\end{center}

\section*{Outline of Supplementary Materials}

\begin{itemize}
\item Proofs of results stated in the main text are provided in \cref{sec:omittedproofs}.
\item Additional experimental results, including coverage plots, are provided in \cref{sec:moreexperiments}.
\item Additional details for all experiments, including specifics on the implementation of the optimization, are provided in \cref{sec:appendix_exp}.
\item The finite-difference IDM algorithm for multivariate-valued evaluations are detailed in \cref{sec:appendix_mv_psi}.
\end{itemize}

\section{Omitted Proofs}\label{sec:omittedproofs}

\subsection{Proof of \cref{thm:iidm}}
\begin{proof}
Let $\theta_0\in\Ncal\subset\Theta$ be the interior neighborhood where the assumed differentiable hold.
Since $\hat\theta_n\to_p\theta_0$, we have that $\Prb{\hat\theta_n\in\Ncal}\to1$.

Define $L_n(\theta)=\sum_{i=1}^n\log f(Z_i;\theta)$ on $\Ncal$.
By assumptions of continuity and existence of envelope, lemma 4.3 of \citetSM{neweymcfaddensm} guarantees that $-\frac1n\nabla^2L_n(\hat\theta_n)\to_p I(\theta_0)$. Since $I(\theta_0)\succ0$, we have by continuous mapping that $\Prb{\nabla^2L_n(\hat\theta_n)\prec0}\to1$.

Consider the event $\hat\theta_n\in\Ncal$ and $\nabla^2L_n(\hat\theta_n)\prec0$, which has probability tending to 1. Define $\Lcal(\lambda,\theta)=\nabla L_n(\theta)+\lambda\nabla\psi(\theta)$ on $\RR\times\Ncal$. Note that $\Lcal(0,\hat\theta_n)=0$ and that the Jacobian of $\Lcal$ with respect to $\theta$ at $(0,\hat\theta_n)$ is $\nabla^2L_n(\hat\theta_n)\prec0$. Therefore, there exists $\tilde\theta_n(\lambda)$ with $\Lcal(\lambda,\tilde\theta_n(\lambda))=0$ and $\frac{\partial\tilde\theta_n}{\partial\lambda}(\lambda)=-(\nabla^2L_n(\hat\theta_n(\lambda;\psi)))^{-1}\nabla\psi(\hat\theta_n(\lambda;\psi))$ for $\lambda\in(-u,u)$ for some $u>0$. Moreover, since $\nabla^2L_n(\hat\theta_n)\prec0$ and by continuous derivatives, there is a $0<u'\leq u$ such that $\tilde\theta_n(\lambda)$ uniquely solves $\Lcal(\lambda,\theta)=0$ for $\lambda\in(-u',u')$ and therefore $\tilde\theta_n(\lambda)=\hat\theta_n(\lambda;\psi)$ for such $\lambda$. 
Finally, chain rule applied to $\psi(\hat\theta_n(\lambda;\psi))$ for $\lambda\in(-u',u')$ gives $\frac{\partial\hat\psi_n}{\partial\lambda}(\lambda)=\nabla\psi(\hat\theta_n(\lambda;\psi))\tr\frac{\partial\hat\theta_n}{\partial\lambda}(\lambda)$.

Applying this at $\lambda=0$, we conclude that with probability tending to 1,
$$n\hat V_n^\IIDM=
\nabla\psi(\hat\theta_n)\tr\prns{-\frac1n\sum_{i=1}^n\nabla^2\log f(Z_i;\hat \theta_n)}^{-1}\nabla\psi(\hat\theta_n).
$$
By continuity, $\nabla\psi(\hat\theta_n)\to_p\nabla\psi(\theta_0)$.
We have also already shown that $-\frac1n\nabla^2L_n(\hat\theta_n)\to_p I(\theta_0)\succ0$.
Therefore, the proof is completed by continuous mapping.
\end{proof}

\subsection{Proof of \cref{thm:fdidm}}

\begin{proof}
The assumptions imply that $\hat\psi_n''(0)$ exists and $n^2\hat\psi_n''(0)$ converges to a constant in probability. Applying Taylor's theorem using the Lagrange form of the remainder, we have that, for some random $\tilde\lambda_n\in[0,\lambda_n]$,
$$
\hat\psi_n(\lambda_n)=\hat\psi_n(0)+\hat\psi_n'(0)\lambda_n+\hat\psi_n''(0)\tilde\lambda_n^2.
$$
Rearranging yields
$$
n\hat V_n^\FDIDM=n\hat V_n^\IIDM+n\hat\psi_n''(0)\tilde\lambda_n^2/\lambda_n.
$$
By \cref{thm:iidm}, $n\hat V_n^\IIDM\to_p V_0$.

Considering the remainder term, we have
$$
\abs{n\hat\psi_n''(0)\tilde\lambda_n^2/\lambda_n}\leq
\frac{\lambda_n}{n}\abs{n^2\hat\psi_n''(0)}=o(1)O_p(1)=o_p(1),
$$
completing the proof.
\end{proof}

\subsection{Proof of \cref{thm:multivar}}

\begin{proof}
From the proof of \cref{eq:iidm}, we know that, with probability tending to 1, 
$\frac{\partial}{\partial\lambda}\hat\theta_n(\lambda;\psi\s j)=-(\nabla^2L_n(\hat\theta_n(\lambda;\psi\s j)))^{-1}\nabla\psi\s j(\hat\theta_n(\lambda;\psi\s j))$ (from the implicit function argument) and $\frac{\partial}{\partial\lambda}\psi\s i(\hat\theta_n(\lambda;\psi\s j))=\nabla\psi\s i(\hat\theta_n(\lambda;\psi\s j))\tr\frac{\partial}{\partial\lambda}\hat\theta_n(\lambda;\psi\s j)$ (from chain rule argument). Combining and applying for each entry $i,j$ yields the first statement. The second statement follows exactly as in \cref{thm:fdidm} by a Taylor expansion.
\end{proof}

\subsection{Proof of \cref{thm:fdidm2}}

\begin{proof}
Define $\dot h(W;\theta)$ as $\nabla h(W;\theta)$ when it exists and $0$ otherwise.
Define $\psi^*(\theta)=\E[h(W;\theta)]$. Further define $$p(\epsilon)=1-\Prb{\text{On $\{\theta:\magd{\theta-\theta_0}\leq\epsilon\}$, $h(W;\theta)$ is twice differentiable in $\theta$ with $\magd{\nabla_\theta^2h(W;\theta)}\leq L$}},$$ so that by assumption $p(\epsilon)=o(1)$.
By continuity of probability, $p(0)=0$ so that $h(W;\theta)$ is almost surely twice boundlessly differentiable at $\theta=\theta_0$. Thus, $\nabla \psi^*(\theta_0)=\E\dot h(W;\theta_0)$ exists. 

Consider any sequence $\theta\to\theta_0$. 
Then, first by triangle inequality and then by Taylor's theorem,
\begin{align*}
\Eb{\abs{\psi(\theta)-\psi(\theta_0)-\nabla \psi^*(\theta_0)\tr(\theta-\theta_0)}}&\leq \Eb{\abs{h(W;\theta)-h(W;\theta_0)-\dot h(W;\theta_0)\tr(\theta-\theta_0)}}
\\&\leq p(\magd{\theta-\theta_0})L\magd{\theta-\theta_0}+\frac12M\magd{\theta-\theta_0}^2
\\&=o(\magd{\theta-\theta_0}).
\end{align*}

Since $\magd{\theta-\theta_0}=O_p(1/\sqrt{n})$, 
$$
\psi(\hat\theta_n)=\psi(\theta_0)+\nabla\psi^*(\theta_0)\tr(\hat\theta_n-\theta_0)+o_p(1/\sqrt{n}).
$$
Therefore, by Slutsky's theorem,
$$
\sqrt{n}(\psi(\hat\theta_n)-\psi(\theta_0))\rightsquigarrow\mathcal N(0,\,\nabla\psi^*(\theta_0)\tr I^{-1}\nabla\psi^*(\theta_0)).
$$
Then if, hypothetically, we apply FDIDM with the unknown $\psi^*(\theta)$ to obtain $\tilde V_n^\FDIDM=(\psi^*(\hat\theta_n(\lambda;\psi^*))-\psi^*(\hat\theta_n))/\lambda$, then, by \cref{thm:fdidm}, we have that 
$$
\frac{\psi(\hat\theta_n)-\psi(\theta_0)}{\sqrt{\tilde V_n^\FDIDM}}\rightsquigarrow\Ncal(0,1).
$$

To complete the proof, we next argue that $\hat V_n^\FDIDM-\tilde V_n^\FDIDM\to_p0$, that is, the difference vanishes between doing hypothetical FDIDM with the unknown $\psi^*$ and actual FDIDM with the known $\psi$.
Let $\theta_0\in\Ncal\subset\operatorname{Interior}(\Theta)$ denote the interior neighborhood where all desirable differentiabilities hold. Our assumptions yield $\Prb*{\hat\theta_n\in\Ncal,\hat\theta_n(\lambda;\psi)\in\Ncal,\hat\theta_n(\lambda;\psi^*)\in\Ncal}\to1$ and $\sup_{\theta\in\Ncal}\abs{\psi(\theta)-\psi^*(\theta)}\to_p0$ per lemma 4.3 of \citetSM{neweymcfaddensm}. Then, on the event that $\hat\theta_n\in\Ncal,\hat\theta_n(\lambda;\psi)\in\Ncal,\hat\theta_n(\lambda;\psi^*)\in\Ncal$, we have
\begin{align*}
\abs{\hat V_n^\FDIDM-\tilde V_n^\FDIDM}&=
\lambda^{-1}\abs{\psi(\hat\theta_n(\lambda;\psi))-\psi(\hat\theta_n)-\psi^*(\hat\theta_n(\lambda;\psi^*))+\psi^*(\hat\theta_n)}\\&\leq 2\lambda^{-1}\sup_{\theta\in\Ncal}\abs{\psi(\theta)-\psi^*(\theta)},
\end{align*}
yielding the statement. 
\end{proof}

\section{Further Experiments}\label{sec:moreexperiments}
\labelsec{further_exp}

Fig.~\ref{fig:cubic_fit} shows the fitted mean and covariance on a single draw of the quadratic dataset. 
With the inputs more dispersed (than the $\sin$ example) 
we see that all methods are in closer agreement. 
Fig.~\ref{fig:coverage} gives the 95\% coverage of the methods as a function of input. 
The quadratic function has close to 95\% coverage across the inputs for all methods. 
The $\sin$ example restricts the set of available inputs (even on resample) 
and uses a neural network on a small dataset (N=160), 
both of which make it particularly challenging for coverage. 
There is some agreement among 
the methods that use the neural network (IDM and delta method). 
The GP-Matern52 comprises a different model of the mean and covariance 
and is included for reference. All of the methods are at least 20 percentage points away from the true coverage across the inputs.

\begin{figure*}[h!]%
\centering%
\begin{subfigure}[b]{0.45\textwidth}%
\includegraphics[height=3.5cm]{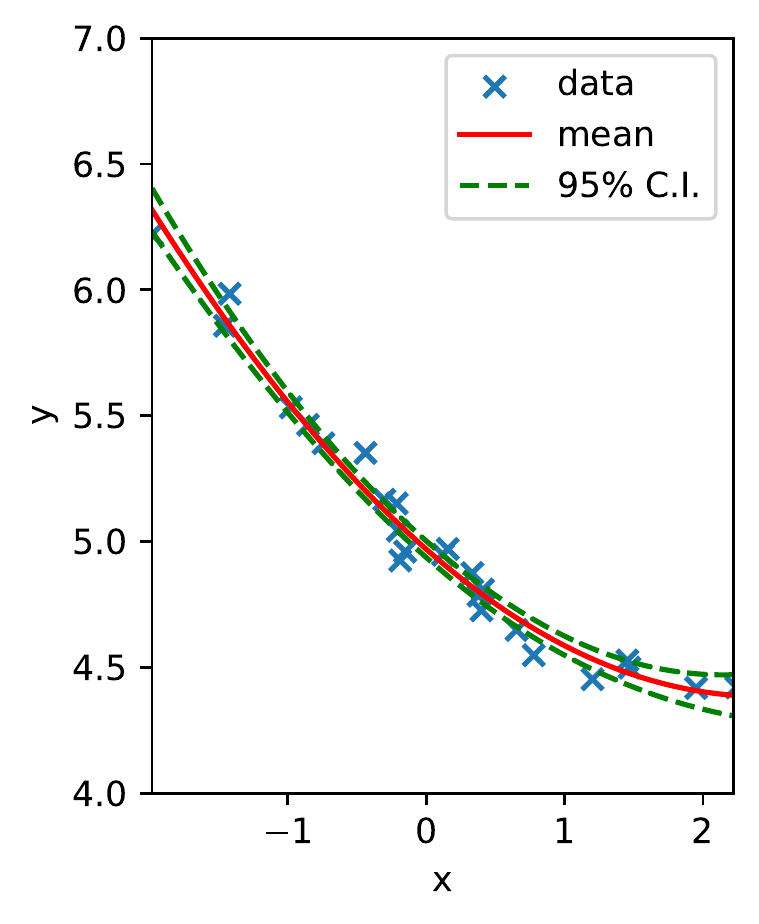}\includegraphics[height=3.5cm]{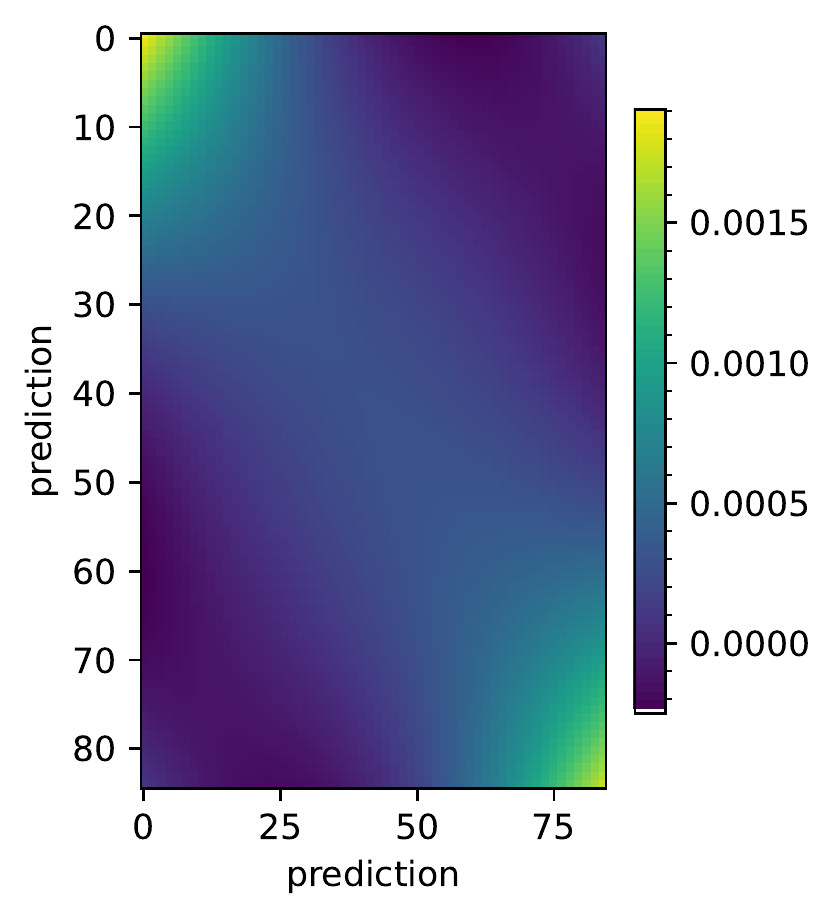}%
\vspace{-2ex}\caption{\capsize \namep (this paper)}%
\end{subfigure}%
\begin{subfigure}[b]{0.45\textwidth}%
\includegraphics[height=3.5cm]{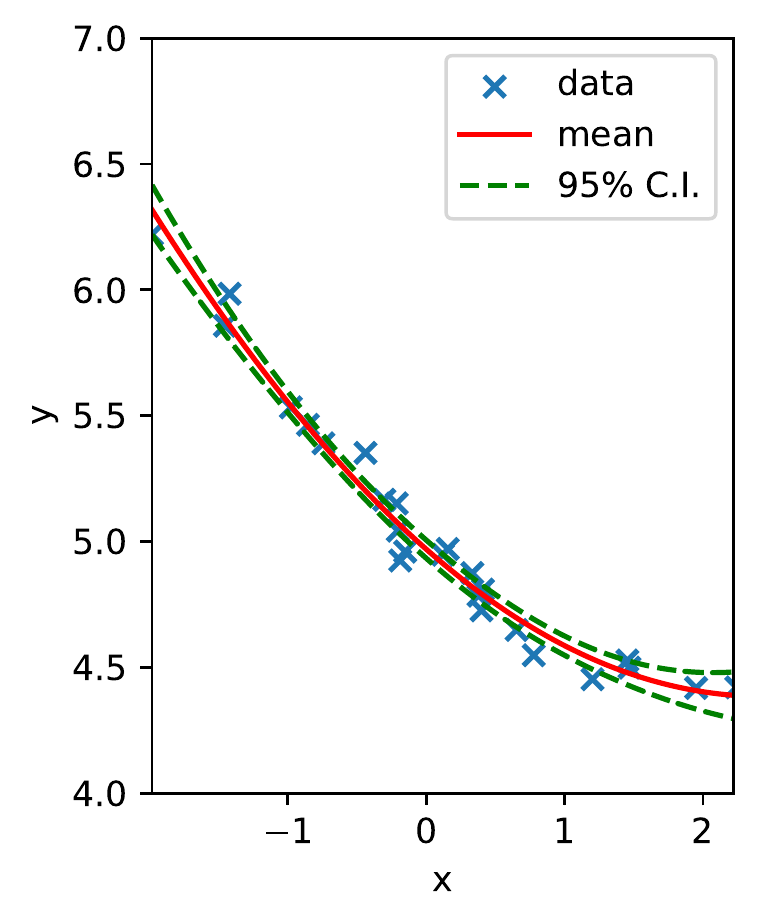}\includegraphics[height=3.5cm]{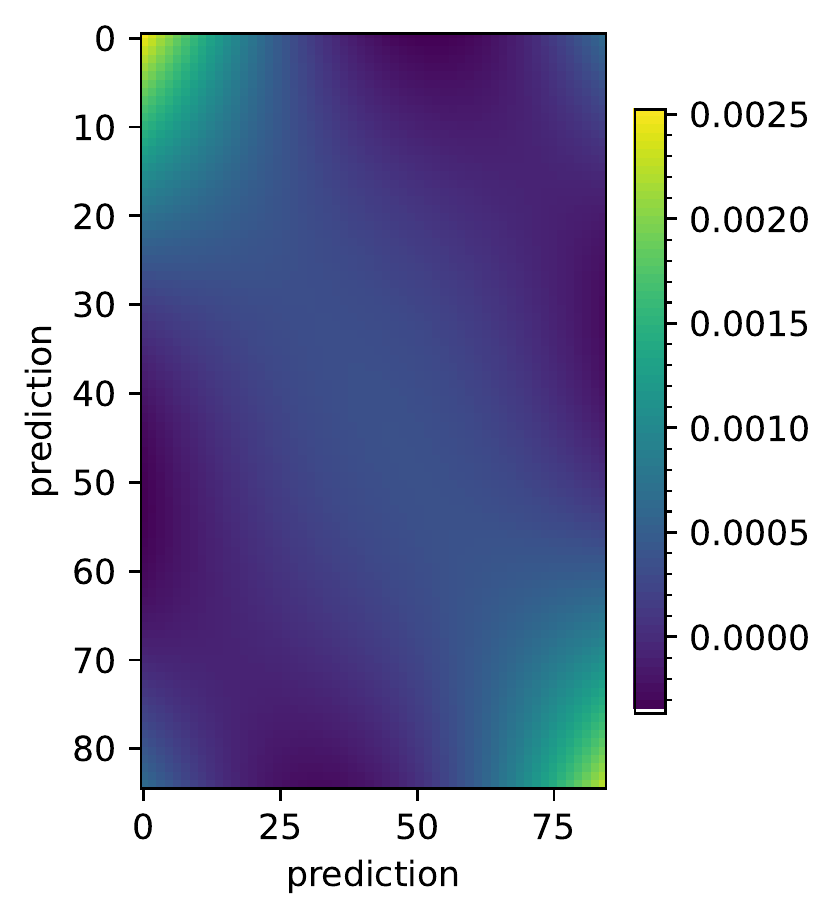}%
\vspace{-2ex}\caption{\capsize Delta Method}%
\end{subfigure}\\[1ex]%
\begin{subfigure}[b]{0.45\textwidth}%
\includegraphics[height=3.5cm]{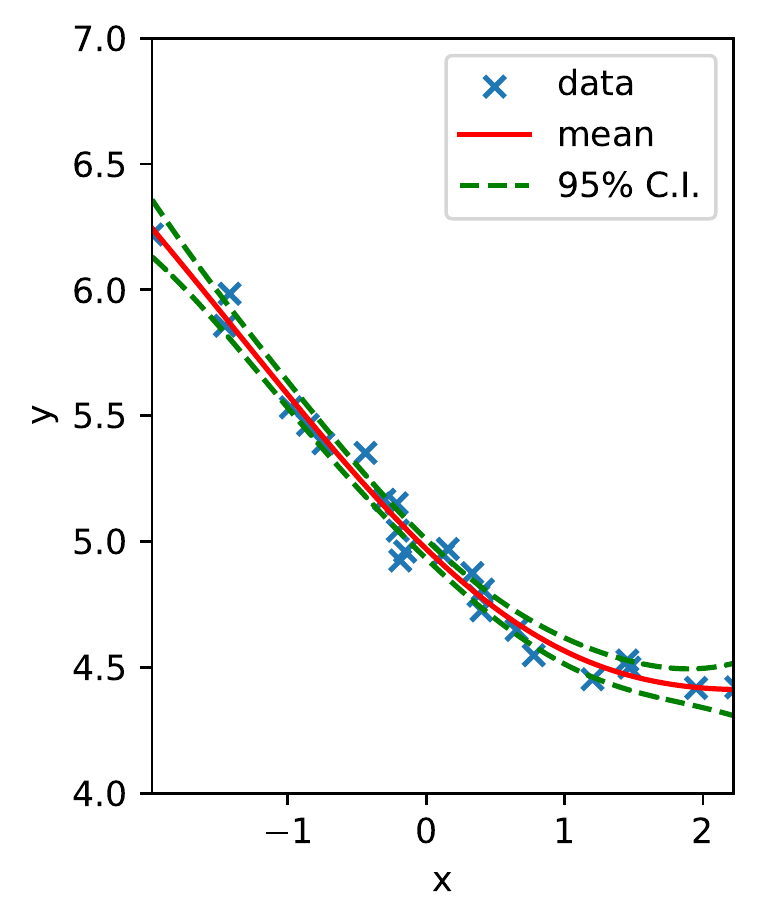}\includegraphics[height=3.5cm]{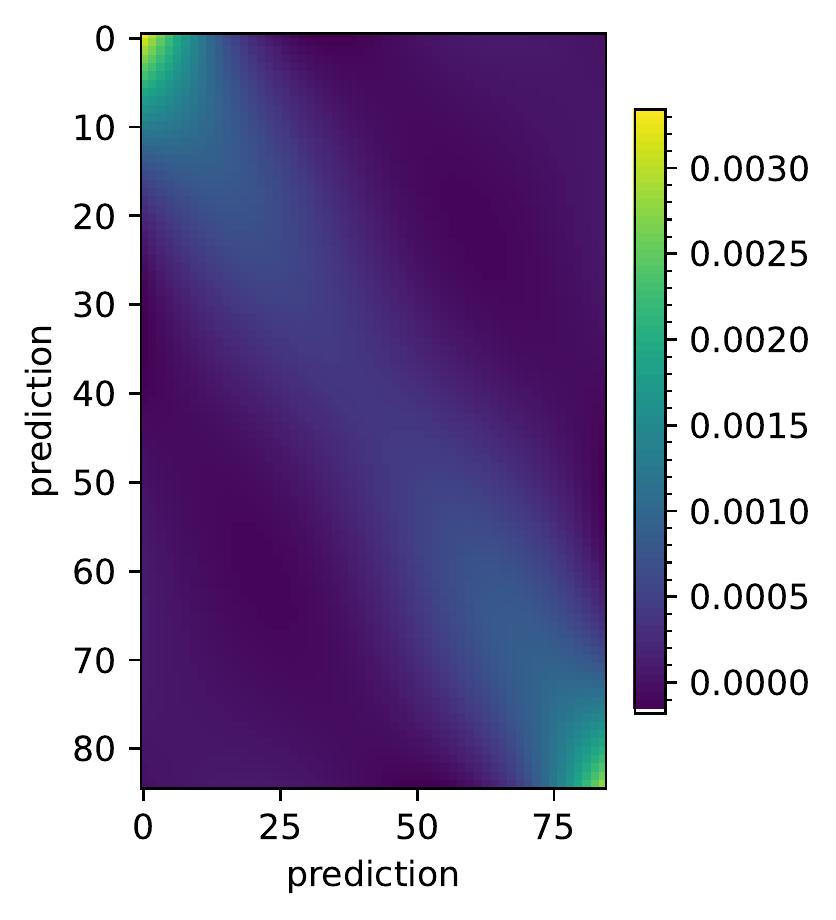}%
\vspace{-2ex}\caption{\capsize GP-Matern52}%
\end{subfigure}%
\begin{subfigure}[b]{0.45\textwidth}%
\includegraphics[height=3.5cm]{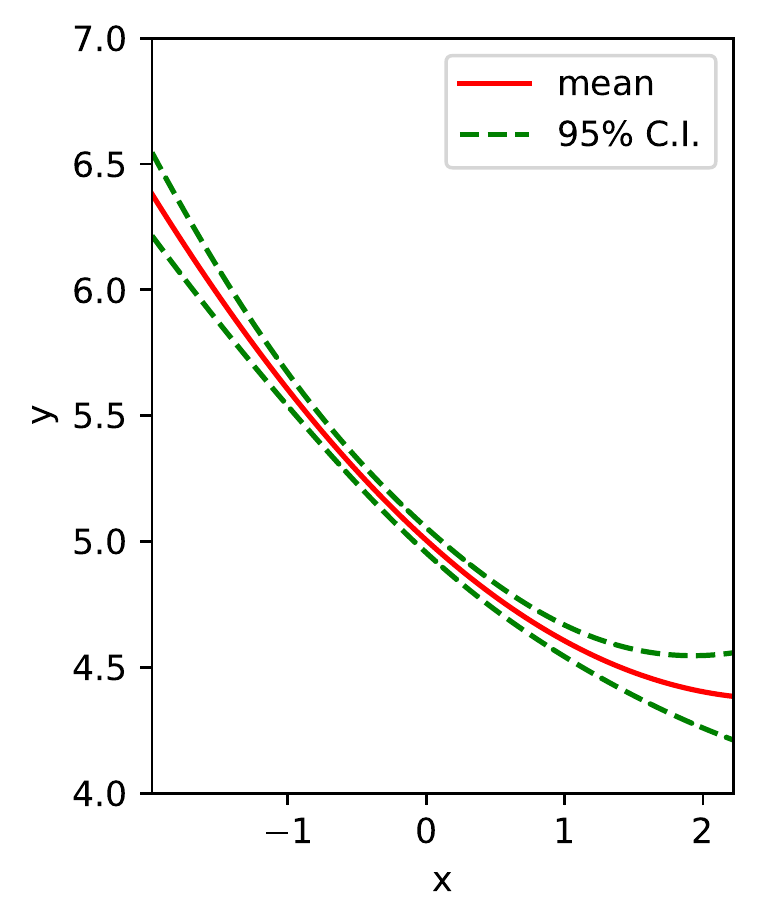}\includegraphics[height=3.5cm]{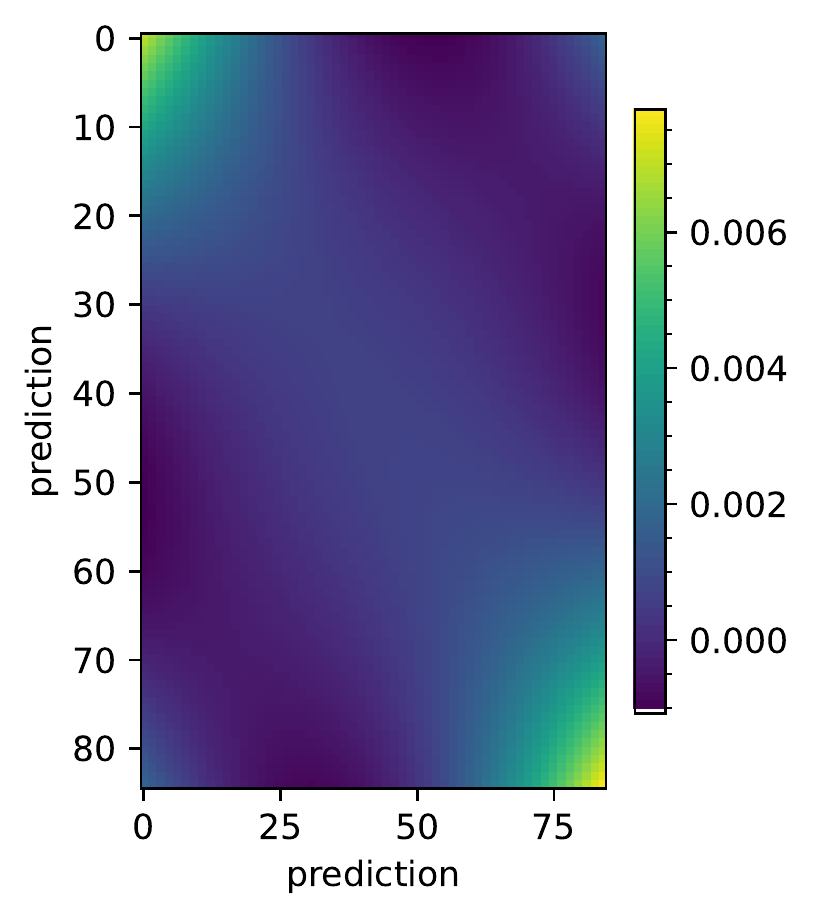}%
\vspace{-2ex}\caption{\capsize True sampling uncertainty by simulation}%
\end{subfigure}%
\caption{Fits along with uncertainty bounds and estimated prediction-covariance matrix for data generated from $y = \frac{1}{10}x^2 - \frac{1}{2}x + 5 + \frac{1}{10}\epsilon$, where $\epsilon \sim \mathcal{N}(0,1)$}%
\labelfig{cubic_fit}%
\end{figure*}

\begin{figure*}[h!]%
\centering%
\begin{subfigure}[b]{0.45\textwidth}%
\includegraphics[width=0.95\linewidth]{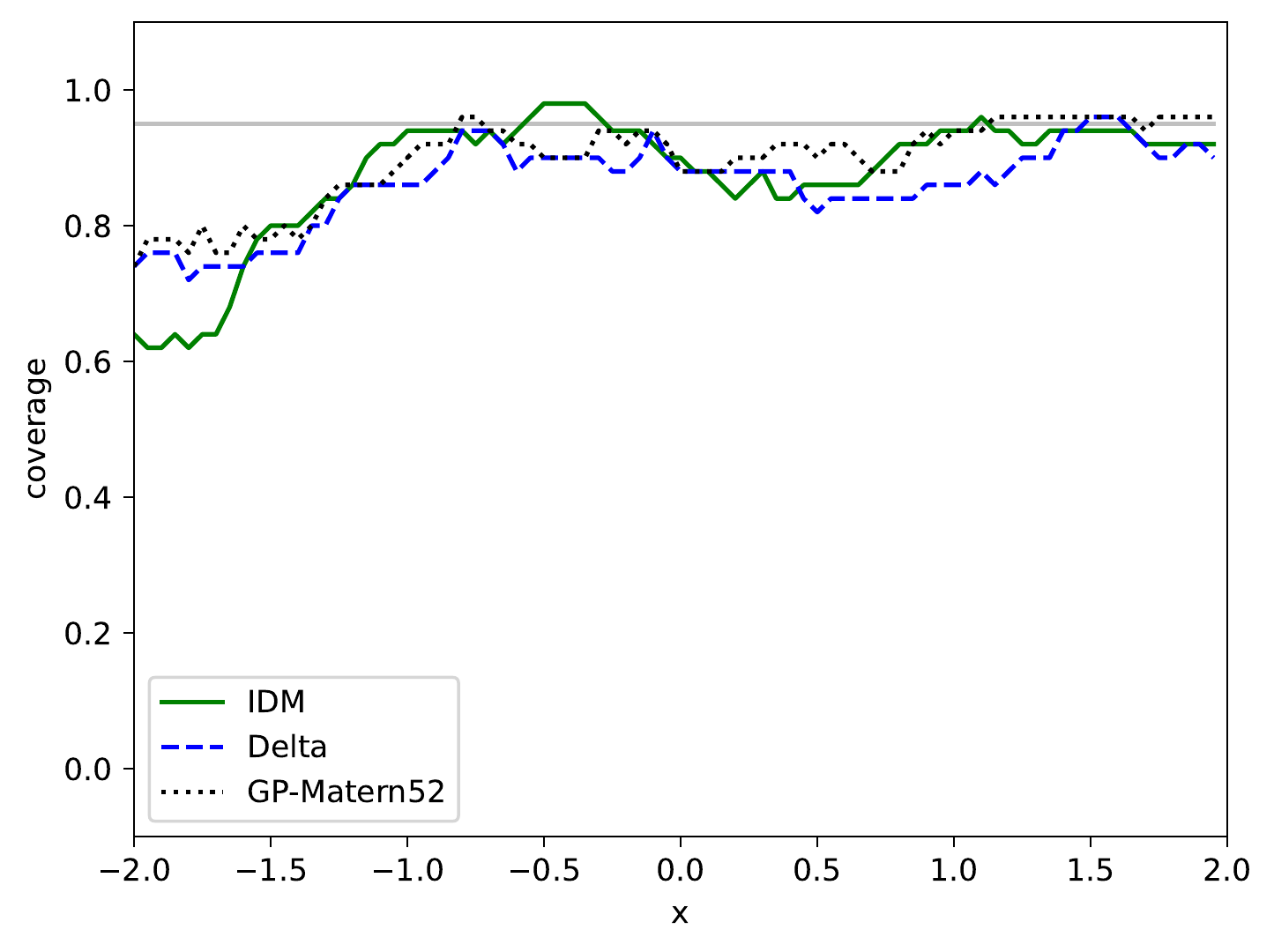}
\caption{\capsize {\bf quadratic} task}%
\end{subfigure}%
\begin{subfigure}[b]{0.45\textwidth}%
\includegraphics[width=0.95\linewidth]{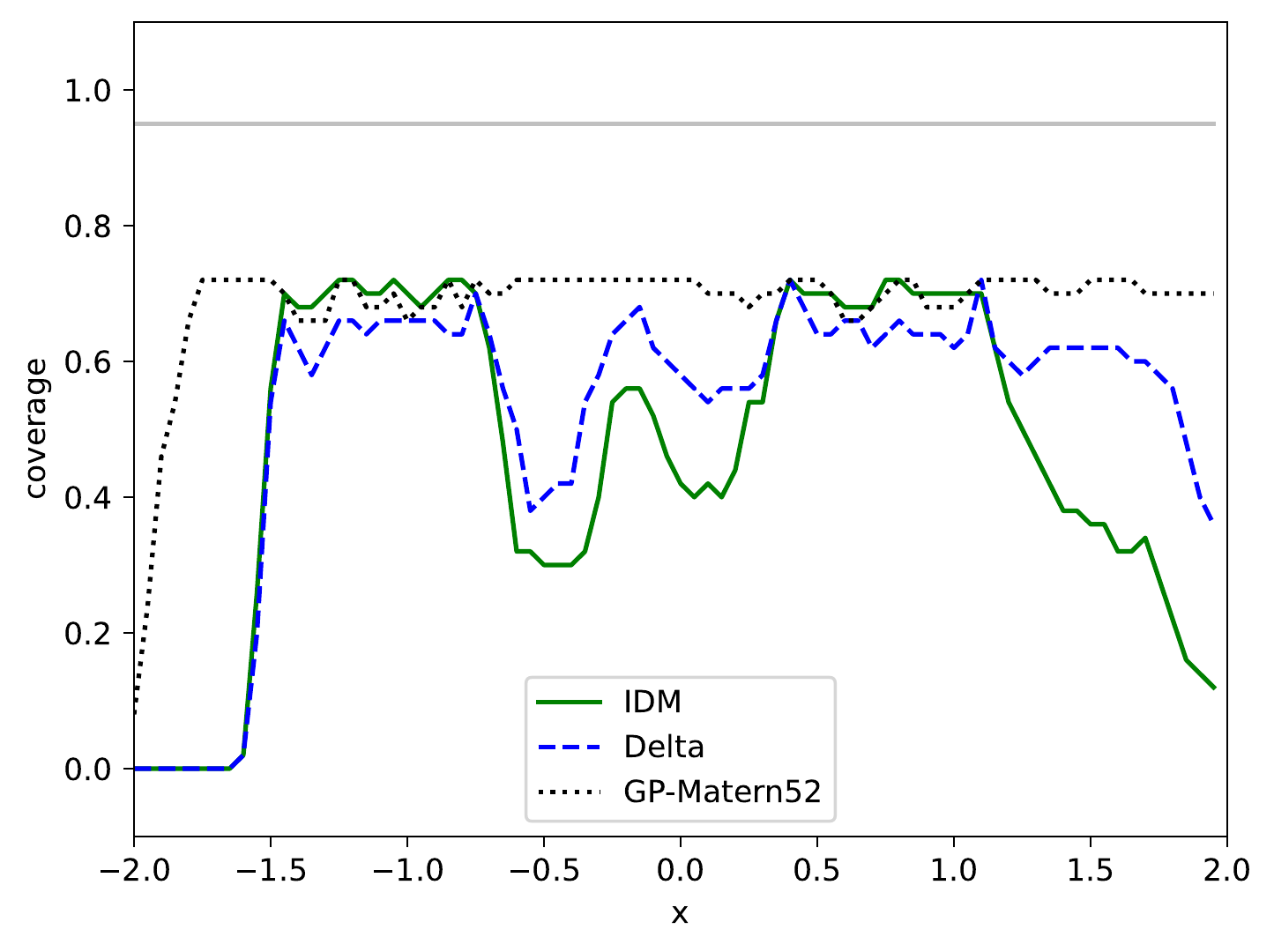}
\caption{\capsize $\boldsymbol{\sin}$ task}%
\end{subfigure} 
\caption{Coverage of 95\% confidence intervals pointwise for prediction at each input.}%
\labelfig{coverage}%
\end{figure*}

\section{Details for Experiments}
\label{sec:appendix_exp}

In this section we provide details for the experimental setup used in the paper. 

\paragraph{Stochastic Gradient Finite-Difference IDM}

The algorithm we propose for combining stochastic gradient ascent 
with finite-difference IDM is given in Alg.~\ref{alg:fdidm_sgd}. 
This algorithm is used in all experiments. 
It estimates $\beta$-confidence intervals 
given a stochastic objective $\tilde{\objective}$ 
that is an unbiased estimate of the true objective $\E[\tilde{\objective}] = \mathcal{L}$ 
and evaluated each iteration $i$ on a new minibatch of the dataset. 
The other inputs are the target evaluation $\psi$, 
scalar width $\lambda$, learning rate schedule $\eta$, 
sample size $S$, and desired confidence $\beta$. 
In practice, we fixed $\lambda$ to be 1\% of the training objective at the MLE, 
though it is also possible to average the IDM output across variable $\lambda$. 

We use TensorFlow v2.1.0 $\texttt{tensorflow.compat.v1}$ to 
compute the gradients of the models in all experiments~\citepSM{tensorflow2015-whitepaper}. 
In experiments with a Gaussian process (GP), we applied the GPFlow package and optimized the marginal likelihood w.r.t. the kernel parameters (type II maximum likelihood)~\citepSM{matthews2017gpflow}.

\begin{algorithm}[t]
\SetAlgoLined
\KwIn{Stochastic objective $\tilde{\objective}$, evaluation $\psi$, scalar $\lambda$, schedule $\eta$, sample size $S$, confidence $\beta$} 
\DontPrintSemicolon
  \SetKwFunction{FMain}{SG-FDIDM}
  \SetKwProg{Fn}{Function}{:}{}
  \Fn{\FMain{$\tilde{\mathcal{L}}$, $\psi$, $\lambda$, $\eta$, $S$, $\beta$}}{
	\tcp{randomly initialize parameters}
        $\hat{\theta}_n \sim \mathrm{Uniform}(-0.1, 0.1)$\;
        \tcp{maximize $\mathcal{L}$ with stochastic gradient ascent}
        $i \gets 0$\;
        \While{$\hat{\theta}_n$ not converged}{
          $\hat{\theta}_n \gets \hat{\theta}_n + \eta(i) \nabla_\theta \tilde{\mathcal{L}}_i$\;
          $i \gets i + 1$
  	}
	\tcp{collect $S$ samples for estimating $\psi(\hat{\theta}_n)$}
	\ForAll{$s \in [0,S)$}{
	$\psi_{0,s} \gets \psi(\hat{\theta}_n)$\;
          $\hat{\theta}_n \gets \hat{\theta}_n + \eta(i) \nabla_\theta \tilde{\mathcal{L}}_i$\;
          $i \gets i + 1$
  	}
	
        \tcp{maximize $\mathcal{L} + \lambda \psi$ with stochastic gradient ascent}
        \While{$\hat{\theta}_n$ not converged}{
          $\hat{\theta}_n \gets \hat{\theta}_n + \eta(i) \nabla_\theta (\tilde{\mathcal{L}}_i + \lambda \psi)$\;
          $i \gets i + 1$
  	}
	\tcp{collect $S$ samples for estimating $\psi(\hat{\theta}_n(\lambda))$}
	\ForAll{$s \in [0,S)$}{
	$\psi_{\lambda,s} \gets \psi(\hat{\theta}_n)$\;
          $\hat{\theta}_n \gets \hat{\theta}_n + \eta(i) \nabla_\theta (\tilde{\mathcal{L}}_i + \lambda \psi)$\;
          $i \gets i + 1$
  	}
	\tcp{take average of samples}
	$\bar{\psi}_0 \gets \frac{1}{S} \sum_{s \in [0,S)} \psi_{0, s}$\;
	$\bar{\psi}_\lambda \gets \frac{1}{S} \sum_{s \in [0,S)} \psi_{\lambda, s}$\;
	\KwRet $\bar{\psi}_0 \pm\Phi^{-1}((1+\beta)/2)\sqrt{\frac{1}{\lambda}(\bar{\psi}_\lambda - \bar{\psi}_0)}$  \tcp*{confidence interval for $\psi(\hat{\theta}_n)$}
  }
 \caption{Stochastic gradient finite-difference implicit delta method}
 \labelalg{fdidm_sgd}
\end{algorithm}

\paragraph{1D Synthetic Experiments}

The setup for the synthetic experiments is as follows,
\begin{itemize}
	\item {\bf quadratic}: draw 25 data points $x \sim \mathcal{N}(0, 1)$ and let $y = \frac{1}{10}x^2 - \frac{1}{2}x + 5 + \frac{1}{10}\epsilon$, where $\epsilon \sim \mathcal{N}(0,1)$.  Use constant learning rate $\eta(\cdot) = \frac{1}{100}$.
	\item $\boldsymbol{\sin}$: fix 160 data points evenly spaced in ranges $[-1.5, -0.7)$ and $[0.35, 1.15)$ and let $y = -\sin(3x - \frac{3}{10}) + \frac{1}{10}\epsilon$, where $\epsilon \sim \mathcal{N}(0,1)$. Use constant learning rate $\eta(\cdot) = \frac{1}{200}$ for MLE objective and $\frac{1}{2000}$ for $\psi$-regularized objective.
\end{itemize}

In each case, the objective is evaluated over the entire dataset, such that $\tilde{\mathcal{L}} = \mathcal{L}$ and $S = 1$. 
In accordance with a univariate Gaussian density over the prediction, 
the network is trained to minimize half the sum of squared errors 
and the resulting interval is divided by the root mean squared error after convergence. 
The visualized covariance matrices were 
projecting to ensure positive semi-definiteness. 
This was done by averaging the matrix with its transpose, 
finding the eigendecomposition of the resulting matrix, 
then reconstructing the matrix with only the eigenvectors corresponding to positive eigenvalues.

\paragraph{Benchmark Classification Experiments}

For all classification datasets, the stochastic objective $\tilde{\mathcal{L}}$ is 
defined over minibatches of size $B=128$. 
$\tilde{\mathcal{L}}_i$ is selected by 
stepping repeatedly through the randomly shuffled dataset in blocks of size $B$. 
RMSProp with an initial learning rate of $0.01$ is used for optimization.\footnote{\url{http://www.cs.toronto.edu/~tijmen/csc321/slides/lecture_slides_lec6.pdf}} 
$\psi$ is evaluated over a held-out validation dataset, 
a randomly selected 20\% subset of the data 
for the benchmarks except $\texttt{mnist}$ 
which is evaluated on the standard 10k test split of the 60k-10k partition. 

\paragraph{KL Down-Weighting Experiment}

We use the standard 60k $\texttt{mnist}$ train dataset.  
$\psi$ is defined as the total negative reconstruction loss on 
a random 1k subset of the training data. 
The Adam~\citepSM{kingma2014adam} optimizer is applied to the (possibly) regularized VAE objective 
with initial learning rate $0.001$. 
The experiment adapted the VAE implementation in TensorFlow available at \url{https://github.com/wiseodd/generative-models}.

\section{Finite-Difference IDM for Multivariate $\psi$}
\labelsec{appendix_mv_psi}

In this section we present the multivariate algorithm for finite-difference IDM (FDIDM). 
In Alg.~\ref{alg:fdidm2}, a multivariate evaluation function $\psi$ is taken as input, 
along with a learning objective $\mathcal{L}$ and scalar width $\lambda$ (as in the univariate case). 
The algorithm optimizes the original MLE objective, 
then optimizes $K$ regularized MLE objectives, each with a different dimension of $\psi$. 
To obtain the estimated $K \times K$ covariance matrix $\hat{\Sigma}$ of $\psi$, 
finite differences are calculated by crossing the output dimension 
of $\psi$ corresponding to the row of $\hat{\Sigma}$ 
with the optimized $\psi$-regularized objective corresponding to the column of $\hat{\Sigma}$.

\begin{algorithm}[t]
\SetAlgoLined
\KwIn{Learning objective $\objective$, multivariate evaluation $\psi$, scalar $\lambda$} 
\DontPrintSemicolon
  \SetKwFunction{FMain}{MV-FDIDM}
  \SetKwProg{Fn}{Function}{:}{}
  \Fn{\FMain{$\mathcal{L}$, $\psi$, $\lambda$}}{
        $\hat{\theta}_n \gets \arg \max_\theta \objective(\theta)$ \tcp*{optimize learning objective}
    $K \gets \mathrm{dim}(\psi(\theta))$\;
        \ForAll{$k = 1,\dots,K$}{
          $\hat{\theta}_n^{(k)}(\lambda) \gets \arg \max_\theta \objective(\theta) + \lambda \psi_k(\theta)$ \tcp*{optimize $\psi_k$-regularized objective}
  }
        \KwRet $\left[ \frac{1}{\lambda}(\psi_j(\hat{\theta}_n^{(k)}(\lambda)) - \psi_j(\hat{\theta}_n)) \right]_{j=1:K, k=1:K}$ \tcp*{$K\times{K}$ covariance matrix of $\hat{\psi}_n$}
  }
 \caption{Multivariate finite-difference implicit delta method}
 \labelalg{fdidm2}
\end{algorithm}

\bibliographystyleSM{plainnat}
\bibliographySM{lit}

\end{document}